\documentclass{article}
\usepackage{ijcai22}

\usepackage{times}
\usepackage{soul}
\usepackage{url}
\usepackage[hidelinks]{hyperref}
\usepackage[utf8]{inputenc}
\usepackage[small]{caption}
\usepackage{graphicx}
\usepackage{amsmath}
\usepackage{amsthm}
\usepackage{booktabs}
\usepackage{algorithm}
\usepackage{algorithmic}
\usepackage{amssymb}
\usepackage{amsmath}
\usepackage{multirow}
\usepackage{booktabs}
\usepackage{caption}
\usepackage{float} 
\usepackage{subcaption}
\usepackage{stfloats}

\title{Attention Based Spatial-Temporal Graph Convolutional Recurrent Networks\\for Traffic Forecasting}

\author{
Haiyang Liu$^1$
\and
Chunjiang Zhu$^2$
\and
Detian Zhang$^1$\thanks{Corresponding author: detian@suda.edu.cn}\And
Qing Li$^3$
\affiliations
$^1$Institute of Artificial Intelligence, Department of Computer Science and Technology, Soochow University, Suzhou, China.\\
$^2$Department of Computer Science, University of North Carolina at Greensboro, Greensboro, NC, USA.\\
$^3$Department of Computing, The Hong Kong Polytechnic University, Hong Kong, China.\\
\emails
20215227052@stu.suda.edu.cn,
chunjiang.zhu@uncg.edu,
detian@suda.edu.cn,
qing-prof.li@polyu.edu.hk 
}

\begin{document}

\maketitle

\begin{abstract}
   Traffic forecasting is one of the most fundamental problems in transportation science and artificial intelligence. The key challenge is to effectively model complex spatial-temporal dependencies and correlations in modern traffic data. Existing methods, however, cannot accurately model both long-term and short-term temporal correlations simultaneously, limiting their expressive power on complex spatial-temporal patterns. In this paper, we propose a novel spatial-temporal neural network framework: Attention-based Spatial-Temporal Graph Convolutional Recurrent Network (\emph{ASTGCRN}), which consists of a graph convolutional recurrent module (\emph{GCRN}) and a global attention module. In particular, \emph{GCRN} integrates gated recurrent units and adaptive graph convolutional networks for dynamically learning graph structures and capturing spatial dependencies and local temporal relationships. To effectively extract global temporal dependencies, we design a temporal attention layer and implement it as three independent modules based on multi-head self-attention, transformer, and informer respectively. Extensive experiments on five real traffic datasets have demonstrated the excellent predictive performance of all our three models with all their average \emph{MAE}, \emph{RMSE} and \emph{MAPE} across the test datasets lower than the baseline methods.
\end{abstract}

\section{Introduction}
\noindent With the development of urbanization, the diversification of transportation modes and the increasing number of transportation vehicles (cabs, electric vehicles, shared bicycles, \emph{etc}.) have put tremendous pressure on urban transportation systems, which has led to large-scale traffic congestions that have become a common phenomenon. Traffic congestion has brought serious economic and environmental impacts to cities in various countries, and early intervention in traffic systems based on traffic forecasting is one of the effective ways to alleviate traffic congestion. By accurately predicting future traffic conditions, it provides a reference basis for urban traffic managers to make proper decisions in advance and improve traffic efficiency.

Traffic forecasting is challenging since traffic data are complex, highly dynamic, and correlated in both spatial and temporal dimensions. Traffic congestion on one road segment can impact traffic flow on spatially close road segments, and traffic conditions (\emph{e.g.}, traffic flow or speed) on the same road segment at different time points can have significant fluctuations. How to fully capture the spatial-temporal dependencies of modern traffic data and accurately predict future information is an important problem in traffic forecasting. 

Recently, deep learning has dominated the field of traffic forecasting due to their capability to model complex non-linear patterns in traffic data. Many works used different deep learning networks to model dynamic local and global spatial-temporal dependencies and achieve promising prediction performance. On the one hand, they often used Recurrent Neural Networks (\emph{RNN}) and the variants such as Long Short-Term Memory (\emph{LSTM}) \cite{LSTM1997long} and Gated Recurrent Units (\emph{GRU}) \cite{GRU2014properties} for \emph{temporal} dependency modeling \cite{li2018dcrnn_traffic,bai2019passenger,bai2020adaptive,chen2021z}. Some other studies used Convolutional Neural Networks (\emph{CNN}) \cite{yu2018spatio,wu2020connecting,zhang2020spatio,huang2020lsgcn} or attention mechanisms \cite{guo2019attention,zheng2020gman} to efficiently extract temporal features in traffic data. On the other hand, Graph Convolutional Networks (\emph{GCNs}) \cite{li2018dcrnn_traffic,wu2019graph,song2020spatial,li2021spatial} are widely used to capture complex \emph{spatial} features and dependencies in traffic road network data.

However, \emph{RNN/LSTM/GRU}-based models can only indirectly model sequential temporal dependencies, and their internal cyclic operations make them difficult to capture long-term global dependencies \cite{li2021dynamic}. To capture global information, \emph{CNN}-based models \cite{yu2018spatio,wu2019graph} stack multiple layers of spatial-temporal modules but they may lose local information. The attention mechanism, though effective in capturing global dependencies, is not good at making short-term predictions \cite{zheng2020gman}. Most of the previous attention-based methods \cite{guo2019attention,guo2021learning} also have complex structures and thus high computational complexity. For instance in \cite{guo2021learning}, the prediction of architectures built by multi-layer encoder and decoder, though excellent, is much slower than most prediction models by 1 or 2 orders of magnitude. Furthermore, most of the current \emph{GCN}-based methods \cite{geng2019spatiotemporal} need to pre-define a static graph based on inter-node distances or similarity to capture spatial information. However, the constructed graph needs to satisfy the static assumption of the road network, and cannot effectively capture complex dynamic spatial dependencies. Moreover, it is difficult to adapt graph-structure-based spatial modeling in various spatial-temporal prediction domains without prior knowledge (\emph{e.g.}, inter-node distances). Therefore, effectively capturing dynamic spatial-temporal correlations and fully considering long-term and short-term dependencies are crucial to further improve the prediction performance.

To fully capture local and global spatial-temporal dependencies from traffic data, in this paper we propose a novel spatial-temporal neural network framework: Attention-based Spatial-temporal Graph Convolutional Recurrent Network (\emph{ASTGCRN}). It consists of a graph convolution recurrent module (\emph{GCRN}) and a global attention module. In particular, \emph{GCRN} integrates \emph{GRU} and adaptive graph convolutional networks for dynamically learning graph structures and capturing complex \emph{spatial} dependencies and \emph{local temporal} relationships. To effectively extract \emph{global temporal} dependencies, we design a temporal attention layer and implement it by three modules based on multi-head self-attention, transformer, and informer respectively.

Our main contributions are summarized as follows:
\begin{itemize}
\item We develop a novel spatial-temporal neural network framework, called \emph{ASTGCRN}, that can effectively model dynamic local and global spatial-temporal dependencies in a traffic road network.
\item In \emph{ASTGCRN}, we devise an adaptive graph convolutional network with signals at different depths convoluted and then incorporate it into the \emph{GRU}. The obtained \emph{GRU} with adaptive graph convolution (\emph{GCRN}) can well capture the \emph{dynamic graph structures, spatial features, and local temporal} dependencies.
\item We propose a general attention layer that accepts inputs from \emph{GCRN} at different time points and captures the \emph{global temporal} dependencies. We implement the layer using multi-head self-attention, transformer, and informer to generate three respective models.
\item Extensive experiments have been performed on five real-world traffic datasets to demonstrate the superior performance of all our three models compared with the current state of the art. In particular, our model with transformer improves the average \emph{MAE}, \emph{RMSE}, and \emph{MAPE} (across the tested datasets) by $0.11$, $0.30$, and $0.28$, respectively. We carry out an additional experiment to show the generalizability of our proposed models to other spatial-temporal learning tasks.
\end{itemize}

\section{Related Work}
\subsection{Traffic Forecasting}
\noindent Traffic forecasting originated from univariate time series forecasting. Early statistical methods include Historical Average (\emph{HA}), Vector Auto-Regressive (\emph{VAR}) \cite{zivot2006vector} and Auto-Regressive Integrated Moving Average (\emph{ARIMA}) \cite{lee1999application,williams2003modeling}, with the \emph{ARIMA} family of models the most popular. However, most of these methods are linear, need to satisfy stationary assumptions, and cannot handle complex non-linear spatial-temporal data. 

With the rise of deep learning, it has gradually dominated the field of traffic forecasting by virtue of the ability to capture complex non-linear patterns in spatial-temporal data. RNN-based and CNN-based deep learning methods are the two mainstream directions for modeling temporal dependence. Early RNN-based methods such as \emph{DCRNN} \cite{li2018dcrnn_traffic} used an encoder-decoder architecture with pre-sampling to capture temporal dependencies, but the autoregressive computation is difficult to focus on long-term correlations effectively. Later, the attention mechanism has been used to improve predictive performance \cite{guo2019attention,zheng2020gman,wang2020traffic,guo2021learning}. In CNN-based approaches, the combination of 1-D temporal convolution \emph{TCN} and graph convolution \cite{yu2018spatio,wu2019graph} are commonly used. But CNN-based models require stacking multiple layers to expand the perceptual field. The emergence of \emph{GCNs} has enabled deep learning models to handle non-Euclidean data and capture implicit spatial dependencies, and they have been widely used for spatial data modeling \cite{huang2020lsgcn,song2020spatial}. The static graphs pre-defined according to the distance or similarity between nodes cannot fully reflect the road network information, and cannot make dynamic adjustments during the training process to effectively capture complex spatial dependencies. Current research overcame the limitations of convolutional networks based on static graphs or single graphs, and more adaptive graph or dynamic graph building strategies \cite{wu2020connecting,bai2020adaptive,lan2022dstagnn} were proposed. 

In addition to the above methods, differential equations have also been applied to improve traffic forecasting. \cite{fang2021spatial} capture spatial-temporal dynamics through a tensor-based Ordinary Differential Equation (\emph{ODE}) alternative to graph convolutional networks. \cite{choi2022graph} introduced Neural Control Differential Equations (\emph{NCDEs}) into traffic prediction, which designed two \emph{NCDEs} for temporal processing and spatial processing respectively.
Although there are dense methods for spatial-temporal modeling, most of them lack the capability to focus on both long-term and short-term temporal correlations, which results in the limitations of capturing temporal dependencies and road network dynamics.

\subsection{Graph Convolutional Networks}

\noindent Graph convolution networks can be separated into spectral domain graph convolution and spatial domain graph convolution. In the field of traffic prediction, spectral domain graph convolution has been widely used to capture the spatial correlation between traffic series. \cite{bruna2013spectral} for the first time proposed spectral domain graph convolution based on spectral graph theory. The spatial domain signal is converted to the spectral domain by Fourier transform, and then the convolution result is inverted to the spatial domain after completing the convolution operation. The specific formula is defined as follows:
\begin{equation}\label{eq2}
g_{\theta }\star_{G}x = g(L)x = Ug_{\theta }(\Lambda)U^{\mathbf{T}}x,
\end{equation}
In the equation, $\star_{G}$ denotes the graph convolution operation between the convolution kernel $g_{\theta }$ and the input signal $x$ and $L=D^{-\frac{1}{2}}{\bf{L}}D^{-\frac{1}{2}}=U \Lambda U^{\mathbf{T}} \in \mathbb {R}^{N \times N}$ is the symmetric normalized graph Laplacian matrix, where $D=diag({\textstyle \sum_{j=1}^{N}}A_{1j}, \cdots, {\textstyle \sum_{j=1}^{N}}A_{Nj})\in \mathbb {R}^{N \times N}$ is the diagonal degree matrix and ${\bf{L}}=D-A$ is the graph Laplacian matrix. $U$ is the Fourier basis of $G$ and $\Lambda$ is the diagonal matrix of $\bf{L}$ eigenvalues. However, the eigenvalue decomposition of the Laplacian matrix in Eq. \eqref{eq2} requires expensive computations. For this reason, \cite{defferrard2016convolutional} uses the Chebyshev polynomial to replace the convolution kernel $g_{\theta }$ in the spectral domain:
\begin{equation}\label{eq3}
g_{\theta }\star_{G}x = g(L)x = \sum_{k=0}^{K-1}\beta_{k}T_{k}(\hat{L})x,
\end{equation}
where $[\beta_{0}, \beta_{1}, \dots, \beta_{K-1}]$ are the learnable parameters, and $K \ge 1$ is the number of convolution kernels. ChebNet does not require eigenvalue decomposition of Laplacian matrices, but uses Chebyshev polynomials $T_{0}(\hat{L}) = I_{n}, T_{1}(\hat{L}) = \hat{L}$, and $T_{n+1}(\hat{L}) = 2\hat{L}T_{n}(\hat{L}) - T_{n-1}(\hat{L})$. Here $\hat{L} = \frac{2}{\lambda_{max}}L-I_{n}$ is the scaled Laplacian matrix, where $\lambda_{max}$ is the largest eigenvalue and $I_{n}$ is the identity matrix. When $K=2$ , ChebNet is simplified to GCN \cite{kipf2016semi}.

\subsection{Attention Mechanism}
\noindent Attention mechanism draws on the human selective visual attention mechanism, and its core goal is to select the information that is more critical to the current task from a large amount of information. It has been verified to be very effective in capturing long-range dependencies in numerous tasks \cite{bahdanau2014neural,hu2018squeeze}. Among various attention mechanisms, the Scaled Dot-Product Attention is one of the most widely used methods. As defined below, it calculates the dot product between the query and the key, and divides it by $\sqrt{d}$ (the dimension of the query and the key) to preserve the stability of the gradient.
\begin{equation}\label{eq8}
Att(Q,K,V) = softmax(\frac{QK^{\mathbf{T}}}{\sqrt{d}})V, 
\end{equation}
where $Q\in \mathbb {R}^{T \times d_{q}}, K\in \mathbb {R}^{T \times d_{k}}$, and $V\in \mathbb {R}^{T \times d_{v}}$ are query, key and value, and $d_{q}=d_{k}=d$ and $d_{v}$ are their dimension. Transformer \cite{vaswani2017attention} abandons traditional architectures such as \emph{RNN} or \emph{CNN}, and is based on encoder-decoder, which effectively solves the problem that \emph{RNN} cannot be easily parallelized and \emph{CNN} cannot efficiently capture long-term dependencies. Taking this advantage, Transformer achieves state-of-the-art performance on multiple \emph{NLP} and \emph{CV} tasks \cite{radford2019language,bai2021segatron,lin2021end}. Informer \cite{zhou2021informer} improves the Transformer and is used to improve the prediction problem of long sequences.


\section{Methodology}

\begin{figure}[t]
\centering
\includegraphics[width=.90\columnwidth]{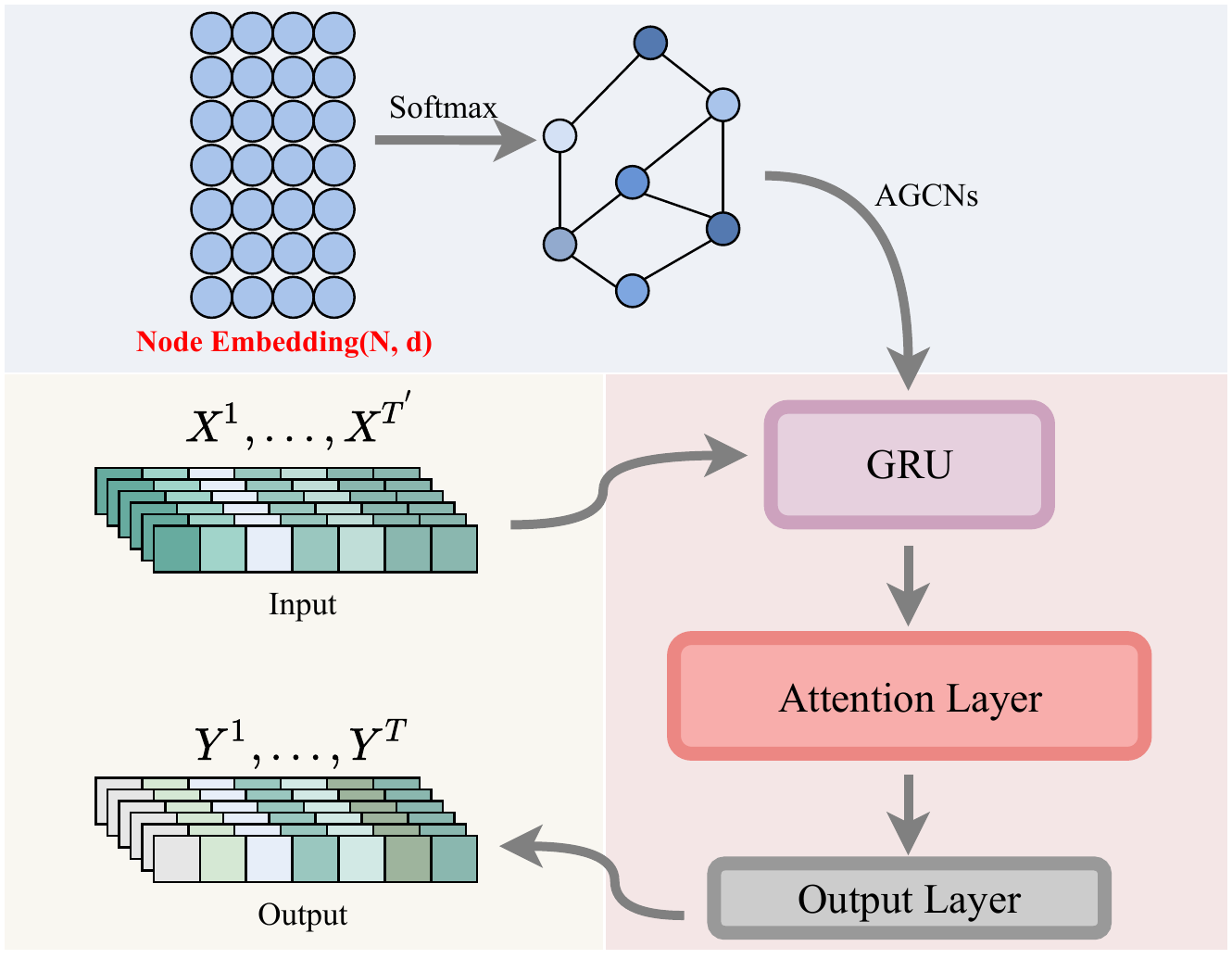} 
\vspace{-0.10in}
\caption{Detailed framework of the \emph{ASTGCRN} model.}
\label{astgcrn}
 \vspace{-0.10in}
\end{figure}

\noindent  In this section, we first give a mathematical definition of the traffic prediction problem, and then detail the two main modules of the \emph{ASTGCRN} framework: \emph{GCRN} and the attention layer. Finally we outline the overall design of this framework.

\vspace{-0.10in}
\subsection{Problem Definition}
\noindent The traffic prediction task can be formulated as a multi-step time series prediction problem that utilizes historical traffic data and prior knowledge of $N$ locations (\emph{e.g.}, traffic sensors) on a road network to predict future traffic conditions. Typically, prior knowledge refers to the road network represented as a graph $G=(V,E,\bf {A})$, where $V$ is a set of $N=|V|$ nodes representing different locations on the road network, $E$ is a set of edges, and $\bf {A} \in \mathbb {R}^{N \times N}$ is the weighted adjacency matrix representing the proximity between nodes (\emph{e.g.}, the road network between nodes). We can formulate the traffic prediction problem as learning a function $F$ to predict the graph signals $Y^{(t+1):(t+T)} \in {\mathbb {R}^{T \times N \times C}}$ of the next $T$ steps based on the past ${T}^{'}$ steps graph signals ${{X}^{(t-{{T}^{'}}+1):t}}\in{{\mathbb{R}}^{{{T}^{'}}\times N\times C}}$ and $G$:
\begin{equation}\label{eq1}
[{{X}^{(t-{{T}^{'}}+1):t}},G]\stackrel{F_{\Theta}}{\longrightarrow}[{{X}^{(t+1):(t+T)}}],
\end{equation}
where $\Theta$ denotes all the learnable parameters in the model.

\vspace{-0.10in}
\subsection{Adaptive Graph Convolution}
\noindent For traffic data in a road network, the dependencies between different nodes may change over time, and the pre-defined graph structure cannot contain complete spatial dependency information. Inspired by the adaptive adjacency matrix \cite{wu2019graph,bai2020adaptive,chen2021z}, we generate $T_{1}(\hat{L})$ in Eq. \eqref{eq3} by randomly initializing a learnable node embedding $E_{\phi }\in \mathbb {R}^{N \times D_{e}}$, where $D_{e}$ denotes the size of the node embedding:
\begin{equation}\label{eq4}
T_{1}(\hat{L}) = \hat{L} = softmax(E_{\phi }\cdot E_{\phi}^{\mathbf{T}})
\end{equation}
To explore the hidden spatial correlations between node domains at different depths, we concatenate $T_{k}(\hat{L})$ at different depths as a tensor $\tilde{T_{\phi }} = [I, T_{1}(\hat{L}), T_{2}(\hat{L}), \dots, T_{K-1}(\hat{L})]^{\mathbf{T}}\in \mathbb {R}^{K \times N \times N}$ and generalize to high-dimensional graph signals $X\in \mathbb {R}^{N \times C_{in}}$. Let $C_{in}$ and $C_{out}$ represent the number of input and output channels, respectively. Then the graph convolution formula in Eq. \eqref{eq3} can be refined as:
\begin{equation}\label{eq5}
g_{\theta }\star_{G}x = g(L)x = \tilde{T_{\phi }}X \Psi,
\end{equation}
where the learnable parameters  $\Psi \in \mathbb {R}^{K \times C_{in} \times C_{out}}$. However, the parameters shared by all nodes have limitations in capturing spatial dependencies \cite{bai2020adaptive}. Instead, we assign independent parameters to each node to get the parameters $\hat{\Psi} \in \mathbb {R}^{N \times K \times C_{in} \times C_{out}}$, which can more effectively capture the hidden information in different nodes. We further avoid overfitting and high spatial complexity problems by matrix factorization. That is to learn two smaller parameters to generate $\hat{\Psi} = E_{\phi}W$, where $E_{\phi}\in \mathbb {R}^{N \times D_{e}}$ is the node embedding dictionary and $W\in \mathbb {R}^{D_{e} \times K \times C_{in} \times C_{out}}$ are the learnable weights. Our adaptive graph convolution formula can be expressed as:
\begin{equation}\label{eq6}
g_{\theta }\star_{G}x = g(L)x = \tilde{T_{\phi }}XE_{\phi}W\in \mathbb {R}^{N \times C_{out}}
\end{equation}

\begin{figure}[t]
\centering
\includegraphics[width=.95\columnwidth]{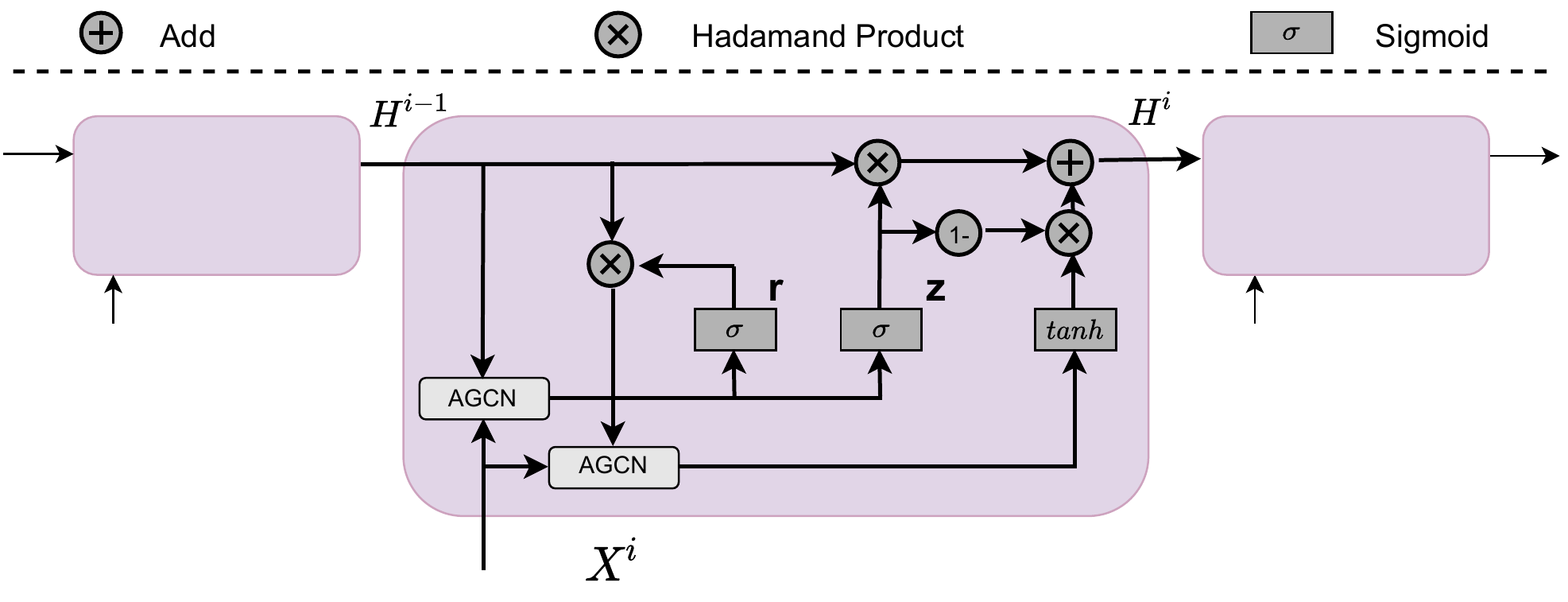} 
\vspace{-0.10in}
\caption{The Graph Convolution Recurrent Module.}
\label{ggru}
 \vspace{-0.10in}
\end{figure}
\vspace{-0.10in}
\subsection{GRU with Adaptive Graph Convolution}
\noindent \emph{GRU} is a simplified version of \emph{LSTM} with multiple \emph{GRUCell} modules and generally provides the same performance as \emph{LSTM} but is significantly faster to compute. To further discover the spatial-temporal correlation between time series, we replace the \emph{MLP} layers in \emph{GRU} with adaptive graph convolution operation, named \emph{GCRN}. The computation of \emph{GCRN} is given as follows:
\begin{equation}
    \begin{split}\label{eq7}
    z^{t} &= \sigma(\tilde{T_{\phi }}[X^{t},h^{t-1}]E_{\phi}W_{z} + E_{\phi}b_{z}), \\
    r^{t} &= \sigma(\tilde{T_{\phi }}[X^{t},h^{t-1}]E_{\phi}W_{r} + E_{\phi}b_{r}), \\
    \tilde{h}^{t} &= tanh(\tilde{T_{\phi }}[X^{t},r^{t} \odot h^{t-1}]E_{\phi}W_{\tilde{h}} + E_{\phi}b_{\tilde{h}}), \\
    h^{t} &= z^{t}\odot h^{t-1} + (1-z^{t})\odot \tilde{h}^t,
    \end{split}
\end{equation}
where $W_{z}, W_{r}, W_{\tilde{h}}, b_{z}, b_{r}$ and $b_{\tilde{h}}$ are learnable parameters, $\sigma$ and $tanh$ are two activation functions, \emph{i.e.}, the Sigmoid function and the Tanh function. The $[X^{t},h^{t-1}]$ and $h^{t}$ are the input and output at time step $t$, respectively. The network architecture of \emph{GCRN} is plotted in Figure \ref{ggru}.

\vspace{-0.10in}
\subsection{The Attention Layer}
\noindent \emph{GCRN} can effectively capture sequential dependencies, but its structural characteristics limit its ability to capture long-distance temporal information. For the traffic prediction task, the global temporal dependence clearly has a significant impact on the learning performance. Self-attention directly connects two time steps through dot product calculation, which greatly shortens the distance between long-distance dependent features, and improves the parallelization of computation, making it easier to capture long-term dependencies in traffic data. Therefore, we propose three independent modules for the self-attention mechanism, namely, the multi-headed self-attention module, the transformer module, and the informer module, in order to directly capture global temporal dependencies. In the following three subsections, we will explain these three modules in detail.

\begin{figure}[t]
\centering
\includegraphics[width=.95\columnwidth]{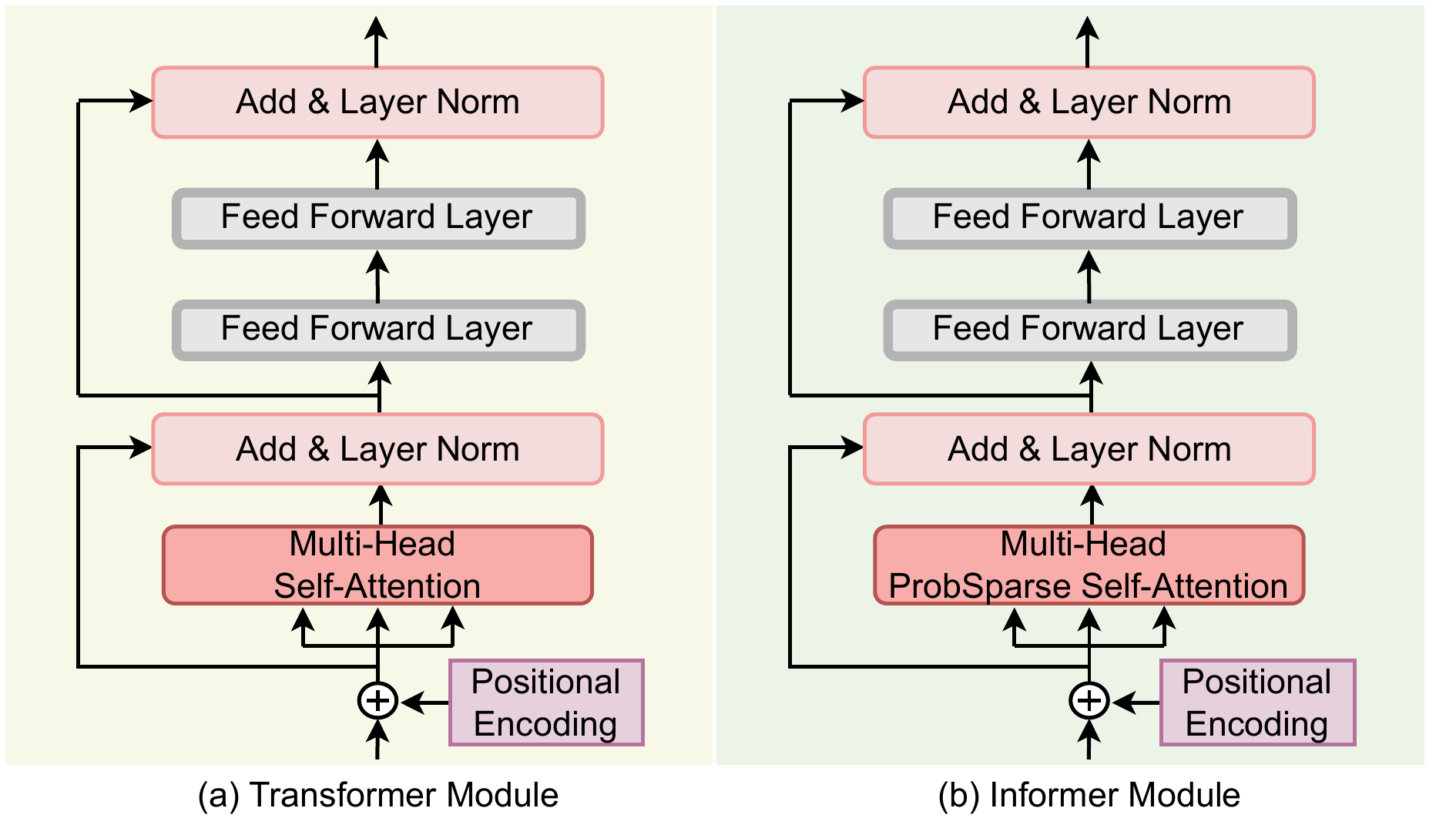} 
\vspace{-0.10in}
\caption{The Attention Layer.}
\label{att}
 \vspace{-0.10in}
\end{figure}

{\noindent \bf Multi-Head Self-Attention Module.}
\noindent Multi-head attention is to learn the dependencies of different patterns in parallel with multiple sets of queries, keys and values (where each set is regarded as an attention head), and then concatenate the learned multiple relationships as the output. We use a self-attentive mechanism to construct the multi-headed attention module. Specifically, $Q_{o}=H_{o}W_{q}$, $K_{o}=H_{o}W_{k}$ and $V_{o}=H_{o}W_{v}$ are derived from the same matrix $H_o$ by linear transformation. Here
$H_{o}\in \mathbb {R}^{N \times T \times C_{out}}$ is the output result of the \emph{GCRN} module, and $W_{q}\in \mathbb {R}^{C_{out} \times d_{q}}$, $W_{k}\in \mathbb {R}^{C_{out} \times d_{k}}$ and $W_{v}\in \mathbb {R}^{C_{out} \times d_{v}}$ are the learnable parameters of the linear projection. For multi-head self-attention mechanism, the formula can be stated as:
\begin{equation}
    \begin{split}\label{eq9}
    MHSelfAtt &= Concat(head_{1}, \dots, head_{h})W_o, \\
    \text{where}\ head_{i} &= Att(Q_{o}, K_{o}, V_{o}) \\
    &= softmax(\frac{Q_{o}{K_{o}}^{\mathbf{T}}}{\sqrt{d}})V_{o}.
    \end{split}
\end{equation}

{\noindent \bf Transformer Module.}
\noindent The Transformer module (see Figure \ref{att}(a)) contains a multi-head self-attention layer and two feed-forward neural networks. For self-attention, each position of the input sequence is equally inner product, which results in the loss of sequential information. We use a fixed position encoding \cite{vaswani2017attention} to address this flaw:
\begin{equation}
    \begin{split}\label{eq10}
    PE_{t}(2c) &= sin(t/1000^{{2c}/{C_{out}}}), \\
    PE_{t}(2c+1) &= cos(t/1000^{{2c}/{C_{out}}}),
    \end{split}
\end{equation}
where $t$ is the relative position of each sequence (time step) of the input and $c$ represents the dimension. In order to better identify the relative positional relationship between sequences, $H_o$ and position encoding are combined to generate $H^{'}_{o}\in \mathbb {R}^{N \times T \times C_{out}}$:
\begin{equation}\label{eq11}
H^{'}_{o}[:,t,:]=H_{o}[:,t,:]+PE_{t}
\end{equation}
After combining the location encoding, $H^{'}_{o}$ is fed as input into the multi-headed self-attentive layer for remote relationship capture, and then the output state is passed to the two fully connected layers. Layer normalization and residual connectivity are used in both sub-layers. Finally, Transformer module outputs the result $H_{a}\in \mathbb {R}^{N \times T \times C_{out}}$.

\smallskip
{\noindent \bf Informer Module.}
\noindent According to Eq. \eqref{eq9}, the traditional self-attention mechanism requires two dot products and $O(T^2)$ space complexity. The sequence lengths of queries and keys are equal in self-attention computation, \emph{i.e.}, $T_{q}=T_{k}=T$. After finding that most of the dot products have minimal attention and the main attention is focused on only a few dot products, \cite{zhou2021informer} proposed ProbSparse self-attention. ProbSparse self-attention selects only the more important queries to reduce the computational complexity, \emph{i.e.}, by measuring the dilution of the queries and then selecting only the top-$u$ queries with $u=c\cdot lnT$ for constant $c$. The query dilution evaluation formula is as follows:
\begin{equation}\label{eq12}
\bar{M}\left({q}_{i}, {K_{o}}\right) = \max _{j}\left\{\frac{{q}_{i} {k}_{j}^{\mathbf{T}}}{\sqrt{d}}\right\}-\frac{1}{T} \sum_{j=1}^{T} \frac{{q}_{i} {k}_{j}^{\mathbf{T}}}{\sqrt{d}}
\end{equation}
where ${q}_{i}$ and ${k}_{i}$ represent the i-th row in $Q_{o}$ and $K_{o}$ respectively. To compute $\bar{M}$, only $U=T\ln T$ dot product pairs are randomly selected, and the other pairs are filled with zeros. In this way, the time and space complexity are reduced to only $O(T\ln T)$. Therefore, we construct a new module called Informer module (see Figure \ref{att}(b)) by using ProbSparse self-attention to replace the normal self-attention mechanism of Transformer module. It selects top-$u$ query according to $\bar{M}$ to generate sparse matrix $Q_{o}^{spa}$. Then the Multi-head ProbSparse self-attention can be expressed as:
\begin{equation}
    \begin{split}\label{eq13}
    {MHProbSelfAtt} &= Concat(head_{1}, \dots, head_{h})W_o, \\
    \text{where}\ head_{i} &= Att(Q_{o}^{spa}, K_{o}, V_{o}) \\
    &= softmax(\frac{Q_{o}^{spa}{K_{o}}^{\mathbf{T}}}{\sqrt{d}})V_{o}.
    \end{split}
\end{equation}

\vspace{-0.2in}
\subsection{Framework of ASTGCRN}
\noindent As in Figure \ref{astgcrn}, our \emph{ASTGCRN} framework uses \emph{GCRN} and attention mechanisms. First, we take the historical data ${{X}^{(t-{{T}^{'}}+1):t}}\in \mathbb {R}^{T \times N \times C}$ input to the \emph{GCRN} module to produce an output $H_{o}\in \mathbb {R}^{N \times T \times C_{out}}$. Then $H_{o}$ is used as the input of the attention layer to get the output $H_{a}\in \mathbb {R}^{N \times T \times C_{out}}$. Finally, the next T-step of data ${{Y}^{(t+1):(t+T)}}\in \mathbb {R}^{T \times N \times 1}$ is output after two fully connected layers.

We choose the $L1$ loss to formulate the objective function and minimize the training error by back propagation. Specifically, the loss function is defined as follows.
\begin{equation}\label{eq14}
Loss = \frac{1}{T}\sum_{t=0}^{T-1} ({{\hat{Y}}^{t}}-{{Y}^{t}})
\end{equation}
where $\hat{Y}$ is the real traffic data, $Y$ is the predicted data, and $T$ is the total predicted time steps.

\section{Experimental Results}
In this section, we present the results of the extensive experiments we have performed. We start by describing the experimental setups and then discuss the prediction results obtained in the baseline settings. Finally, the ablation study and the effects of hyperparameter tuning are provided.


\begin{table}[t]
\centering
\resizebox{.80\columnwidth}{!}{
	\begin{tabular}{c c c c c}
		\toprule[1.2pt]
		Datasets & Nodes & Samples & Unit & Time Span \\
		\midrule
		PEMSD3 & $358$ & $26,208$ & $5$ mins & $3$ months  \\
		PEMSD4 & $307$ & $16,992$ & $5$ mins & $2$ months  \\
		PEMSD7 & $883$ & $28,224$ & $5$ mins & $3$ months  \\
		PEMSD8 & $170$ & $17,856$ & $5$ mins & $2$ months \\
		PEMSD7(M) & $228$ & $12,672$ & $5$ mins & $2$ months  \\
		DND-US & $53$ & $313$ & $1$ week & $6$ years  \\
		\bottomrule[1.2pt]
	\end{tabular}
	}
	\vspace{-0.1in}
	\caption{Statistics of the tested datasets}
	\label{table1}
 \vspace{-0.25in}
\end{table}

{\noindent \bf Datasets.}
We evaluate the performance of the developed models on five widely used traffic prediction datasets collected by Caltrans Performance Measure System (\emph{PeMS}) \cite{chen2001freeway}, namely \emph{PEMSD3}, \emph{PEMSD4}, \emph{PEMSD7}, \emph{PEMSD8}, and \emph{PEMSD7(M)} \cite{fang2021spatial,choi2022graph}.
The traffic data are aggregated into $5$-minute time intervals, \emph{i.e.}, $288$ data points per day. In addition, we construct a new US natural death dataset \emph{DND-US} to study the generalizability of our method to other spatial-temporal data. It contains weekly natural deaths for $53$ (autonomous) states in the US for the six years from 2014 to 2020. Following existing works \cite{bai2020adaptive}, the Z-score normalization method is adopted to normalize the input data to make the training process more stable. Detailed statistics for the tested datasets are summarized in Table \ref{table1}.

{\noindent \bf Baseline Methods.}
We compare our models with the following baseline methods:
\begin{itemize}
    \item Traditional time series forecasting methods, Historical Average (\emph{HA}), \emph{ARIMA} \cite{williams2003modeling}, \emph{VAR} \cite{zivot2006vector}, and \emph{SVR} \cite{drucker1996support}; 
    \item \emph{RNN}-based models: \emph{FC-LSTM} \cite{sutskever2014sequence}, \emph{DCRNN} \cite{li2018dcrnn_traffic}, \emph{AGCRN} \cite{bai2020adaptive}, and \emph{Z-GCNETs} \cite{chen2021z};
    \item \emph{CNN}-based methods: \emph{STGCN} \cite{yu2018spatio}, \emph{Graph WaveNet} \cite{wu2019graph}, \emph{MSTGCN}, \emph{LSGCN} \cite{huang2020lsgcn}, \emph{STSGCN} \cite{song2020spatial}, and \emph{STFGNN} \cite{li2021spatial}; 
    \item Attention-based models: \emph{ASTGCN(r)} \cite{guo2019attention}, \emph{ASTGNN} \cite{guo2021learning}, and \emph{DSTAGNN} \cite{lan2022dstagnn}; 
    \item Other types of models: \emph{STGODE} \cite{fang2021spatial} and \emph{STG-NCDE} \cite{choi2022graph}. 
\end{itemize}
More details on the above baseline methods can be found in Appendix \ref{apd:baseline}.

{\noindent \bf Experimental Settings.}
All datasets are split into training set, validation set and test set in the ratio of $6$:$2$:$2$. Our model and all baseline methods use the $12$ historical continuous time steps as input to predict the data for the next $12$ continuous time steps.

Our models are implemented based on the Pytorch framework, and all the experiments are performed on an NVIDIA GeForce GTX 1080 TI GPU with 11G memory. The following hyperparameters are configured based on the models' performance on the validation dataset: we train the model with $300$ epochs at a learning rate of $0.003$ using the Adam optimizer \cite{kingma2014adam} and an early stop strategy with a patience number of $15$. The batch size is $64$ for all the datasets except for the \emph{PEMSD7} dataset where the batch size is set to 16. The number of \emph{GCRN} layers is $2$, where the number of hidden units per layer are in $\{32, 64\}$, and the number of convolutional kernels $K = 2$. The weight decay coefficients are varied in $\{0, 0.0001, \cdots, 0.001\}$, and the node embedding dimension $D_{e}$ are varied in $\{2, 4, 6, 8, 10\}$. The details of the hyperparameter tuning are in Appendix \ref{apd:hyperparam}.

Three common prediction metrics, Mean Absolute Error (\emph{MAE}), Root Mean Square Error (\emph{RMSE}), and Mean Absolute Percentage Error (\emph{MAPE}), are used to measure the traffic forecasting performance of the tested methods.
(Their formal definitions are given in Appendix \ref{apd:metrics}.) In the discussions below, we refer to our specific \emph{ASTGCRN} models based on the Multi-head self-attention module, Transformer module, and Informer module as \emph{A-ASTGCRN}, \emph{T-ASTGCRN}, and \emph{I-ASTGCRN}, respectively.

\subsection{Experimental Results}
\begin{table*}[t]
	\renewcommand{\arraystretch}{1.2}
	\centering
	\resizebox{.95\linewidth}{!}{
	\begin{tabular}{c|c c c|c c c |c c c|c c c|c c c}
		\toprule[1.2pt]
		\multirow{2}{*}{Model} & \multicolumn{3}{c|}{PEMSD3} & \multicolumn{3}{c|}{PEMSD4} & \multicolumn{3}{c|}{PEMSD7} & \multicolumn{3}{c}{PEMSD8} & \multicolumn{3}{c}{PEMSD7(M)}\\
		\cline{2-16}
		{}& MAE & RMSE & MAPE & MAE & RMSE & MAPE & MAE & RMSE & MAPE & MAE & RMSE & MAPE & MAE & RMSE & MAPE \\
		\midrule
		HA & 31.58 & 52.39 & 33.78\% & 38.03 & 59.24 & 27.88\% & 45.12 & 65.64 & 24.51\% & 34.86 & 59.24 & 27.88\% & 4.59 & 8.63 & 14.35\% \\
		ARIMA & 35.41 & 47.59 & 33.78\% & 33.73 & 48.80 & 24.18\% & 38.17 & 59.27 & 19.46\% & 31.09 & 44.32 & 22.73\% & 7.27 & 13.20 & 15.38\% \\
		VAR & 23.65 & 38.26 & 24.51\% & 24.54 & 38.61 & 17.24\% & 50.22 & 75.63 & 32.22\% & 19.19 & 29.81 & 13.10\% & 4.25 & 7.61 & 10.28\% \\
		SVR & 20.73 & 34.97 & 20.63\% & 27.23 & 41.82 & 18.95\% & 32.49 & 44.54 & 19.20\% & 22.00 & 33.85 & 14.23\% & 3.33 & 6.63 & 8.53\% \\ \hline
		FC-LSTM & 21.33 & 35.11 & 23.33\% & 26.77 & 40.65 & 18.23\% & 29.98 & 45.94 & 13.20\% & 23.09 & 35.17 & 14.99\% & 4.16 & 7.51 & 10.10\% \\
		DCRNN & 17.99 & 30.31 & 18.34\% & 21.22 & 33.44 & 14.17\% & 25.22 & 38.61 & 11.82\% & 16.82 & 26.36 & 10.92\% & 3.83 & 7.18 & 9.81\% \\
		AGCRN &15.98 & 28.25 & 15.23\% & 19.83 & 32.26 & 12.97\% & 22.37 & 36.55 & 9.12\% & 15.95 & 25.22 & 10.09\% & 2.79 & 5.54 & 7.02\% \\
		Z-GCNETs & 16.64 & 28.15 & 16.39\% & 19.50 & 31.61 & 12.78\% & 21.77 & 35.17 & 9.25\% & 15.76 & 25.11 & 10.01\% & 2.75 & 5.62 & 6.89\% \\ \hline
		STGCN & 17.55 & 30.42 & 17.34\% & 21.16 & 34.89 & 13.83\% & 25.33 & 39.34 & 11.21\% & 17.50 & 27.09 & 11.29\% & 3.86 & 6.79 & 10.06\% \\
		Graph WaveNet & 19.12 & 32.77 & 18.89\% & 24.89 & 39.66 & 17.29\% & 26.39 & 41.50 & 11.97\% & 18.28 & 30.05 & 12.15\% & 3.19 & 6.24 & 8.02\% \\
		MSTGCN & 19.54 & 31.93 & 23.86\% & 23.96 & 37.21 & 14.33\% & 29.00 & 43.73 & 14.30\% & 19.00 & 29.15 & 12.38\% & 3.54 & 6.14 & 9.00\% \\
		LSGCN & 17.94 & 29.85 & 16.98\% & 21.53 & 33.86 & 13.18\% & 27.31 & 41.46 & 11.98\% & 17.73 & 26.76 & 11.20\% & 3.05 & 5.98 & 7.62\% \\
		STSGCN & 17.48 & 29.21 & 16.78\% & 21.19 & 33.65 & 13.90\% & 24.26 & 39.03 & 10.21\% & 17.13 & 26.80 & 10.96\% & 3.01 & 5.93 & 7.55\% \\
		STFGNN & 16.77 & 28.34 & 16.30\% & 20.48 & 32.51 & 16.77\% & 23.46 & 36.60 & 9.21\% & 16.94 & 26.25 & 10.60\% & 2.90 & 5.79 & 7.23\% \\ \hline
		ASTGCN(r) & 17.34 & 29.56 & 17.21\% & 22.93 & 35.22 & 16.56\% & 24.01 & 37.87 & 10.73\% & 18.25 & 28.06 & 11.64\% & 3.14 & 6.18 & 8.12\% \\
            ASTGNN & 15.65 & \underline{25.77} & 15.66\% & \underline{18.73} & \underline{30.71} & 15.56\% & 20.58 & 34.72 & \underline{8.52\%} & \underline{15.00} & \underline{24.59} & \underline{9.49\%} & 2.98 & 6.05 & 7.52\% \\
		DSTAGNN &  \underline{15.57} & 27.21 &  \underline{14.68\%} & 19.30 & 31.46 &  \underline{12.70\%} & 21.42 & 34.51 & 9.01\% & 15.67 &  24.77 & 9.94\% & 2.75 & 5.53 & 6.93\% \\ \hline
		STGODE & 16.50 & 27.84 & 16.69\% & 20.84 & 32.82 & 13.77\% & 22.59 & 37.54 & 10.14\% & 16.81 & 25.97 & 10.62\% & 2.97 & 5.66 & 7.36\% \\
		STG-NCDE &  \underline{15.57} &  27.09 & 15.06\% &  19.21 &  31.09 & 12.76\% &  \underline{20.53} &  \underline{33.84} &  8.80\% &  15.45 & 24.81 &  9.92\% &  \underline{2.68} &  \underline{5.39} &  \underline{6.76\%} \\
		\midrule
		\bf{A-ASTGCRN} & \bf{15.06} & 26.71 & \bf{13.83}\%* & 19.30 & 30.92 & 12.91\% & \bf{20.42}* & \bf{33.81} & 8.54\%* & 15.46 & \bf{24.54} & 9.89\% & \bf{2.66} & \bf{5.36} & \bf{6.72}\% \\
		\bf{I-ASTGCRN} & \bf{15.06} & 26.40 & \bf{13.91}\% & 19.15* & 30.80* & 12.89\% & 20.81 & \bf{33.83} & 8.95\% & 15.26 & \bf{24.53} & 9.65\% & \bf{2.63}* & \bf{5.30}* & \bf{6.60}\%* \\
		\bf{T-ASTGCRN} & \bf{14.90}* & 26.01* & \bf{14.17}\% & 19.21 & 31.05 & \bf{12.67}\%* & \bf{20.53} & \bf{33.75}* & 8.73\% & 15.14* & \bf{24.24}* & 9.63\%* & \bf{2.63}* & \bf{5.32} & \bf{6.66}\% \\
		\bottomrule[1.2pt]
	\end{tabular}
	}
	\vspace{-0.10in}
	\caption{Performance comparison of different models on the tested datasets. Underlined results are the current state of the art among the existing methods. Our three models outperform \emph{almost all} the baseline methods, as shown in bold font. Results marked with $*$ are the best prediction performance we achieve. The prediction of \emph{ASTGNN} is good but much slower than most prediction models by 1 or 2 orders of magnitude!}
	\label{table2}
 \vspace{-0.20in}
\end{table*}

\begin{table}[t]
	\renewcommand{\arraystretch}{1.2}
	\centering
	\resizebox{.90\columnwidth}{!}{
	\begin{tabular}{c|c c c}
		\toprule[1.2pt]
        Model &  MAE & RMSE & MAPE(\%)  \\
		\midrule
		AGCRN & 15.38 (106.2\%) & 25.56 (106.2\%) & 10.89 (105.0\%) \\
            ASTGNN & \underline{14.59 (100.8\%)} & \underline{24.37 (101.2\%)} & 11.35 (108.6\%) \\
		Z-GCNETs & 15.28 (105.5\%) & 25.13 (104.4\%) & 11.06 (106.7\%) \\
		DSTAGNN & 14.94 (103.2\%) & 24.70 (102.6\%) & \underline{10.65 (102.7\%)} \\
		STG-NCDE & 14.69 (101.5\%) & 24.44 (101.5\%) & 10.66 (102.8\%) \\
		\midrule
		\bf{A-ASTGCRN} & \bf{14.58 (100.7\%)} & \bf{24.27 (100.8\%)} & \bf{10.38 (100.1\%)} \\
		\bf{I-ASTGCRN} & \bf{14.58 (100.7\%)} & \bf{24.17 (100.4\%)} & \bf{10.40 (100.3\%)} \\
		\bf{T-ASTGCRN} & \bf{14.48 (100.0\%)} & \bf{24.07 (100.0\%)} & \bf{10.37 (100.0\%)} \\
		\bottomrule[1.2pt]
	\end{tabular}
	}
	\vspace{-0.1in}
	\caption{Mean performance metrics for several competitive methods by averaging over all the tested datasets. The relative performance with respect to our \emph{T-ASTGCRN} method is given in parentheses.}
	\label{table3}
 \vspace{-0.10in}
\end{table}

Table \ref{table2} shows the prediction performance of our three models together with the nineteen baseline methods on the five tested datasets. Remarkably, our three models outperform almost all the baseline methods in prediction on all the datasets and, in some settings, achieve comparable performance with \emph{ASTGNN}. 
Table \ref{cost} lists the training time (s/epoch) and inference time (s/epoch) of our models, as well as several recent and best-performing baselines on the \emph{PEMSD4} dataset. 
It is worth noting that the currently top-performing baselines \emph{ASTGNN}, \emph{DSTAGNN}, and \emph{STG-NCDE} have run-time slower than the proposed models often by 1 to 2 orders of magnitude.

\begin{table}[t]
	\renewcommand{\arraystretch}{1.2}
	\centering
    \small
	\resizebox{.55\columnwidth}{!}{
	\begin{tabular}{c|c c}
		\toprule[1.2pt]
        Model &  Training & Inference \\
		\midrule
		STGODE & 111.77 & 12.19 \\
		Z-GCNETs & 63.34 & 7.40\\
        ASTGNN & 658.41 &  156.56 \\
		DSTAGNN &  242.57 & 14.64  \\
		STG-NCDE & 1318.35 & 93.77 \\
            \midrule
		\bf{A-ASTGCRN} & 45.12 & 5.18\\
		\bf{I-ASTGCRN} & 58.84 & 6.51 \\
		\bf{T-ASTGCRN} & 54.80 & 5.62 \\
		\bottomrule[1.2pt]
	\end{tabular}
	}
	\caption{Computation time on \emph{PEMSD4}.}
	\label{cost}
    \vspace{-0.2in}
\end{table}


The overall prediction results of traditional statistical methods (including \emph{HA}, \emph{ARIMA}, \emph{VR}, and \emph{SVR}) are not satisfactory because of its limited ability to handle non-linear data. Their prediction performance is worse than deep learning methods by large margins. \emph{RNN}-based methods such as \emph{DCRNN}, \emph{AGCRN}, and \emph{Z-GCNRTs} suffer from the limitation of \emph{RNNs} that cannot successfully capture long-term temporal dependence and produce worse results than our methods. \emph{CNN}-based models such as \emph{STGCN}, \emph{Graph WaveNet}, \emph{STSGCN}, \emph{STFGCN}, and \emph{STGODE}, have either worse or comparable performance compared to \emph{RNN}-based methods in our empirical study.  They get the 1-D \emph{CNN} by temporal information, but the size of the convolutional kernel prevents them from capturing the complete long-term temporal correlation. \emph{ASTGCN}, \emph{ASTGNN} and \emph{DSTAGNN} all use the temporal attention module, they all enhance the information capture ability by stacking multi-layer modules, but this also leads to their lack of local feature capture ability and huge training cost. Furthermore, they are applicable to the task without providing prior knowledge. \emph{STG-NCDE} achieves currently best performance in multiple datasets. But their temporal \emph{NCDE} using only the fully connected operation cannot pay full attention to the temporal information. Table \ref{table3} presents the average values of \emph{MAE}, \emph{RMSE}, and \emph{MAPE} across all five datasets for our three models and several top-performing baselines. All our three methods consistently achieve a better average performance while \emph{T-ASTGCRN} obtains the best prediction accuracy on average. This is possibly due to that \emph{T-ASTGCRN} introduces position encoding to retain position information while keeping all attention information generated by dot product computation without discarding any of them. 

To test the generalizability of our proposed models to other spatial-temporal learning tasks, we perform an additional experiment on the \emph{DND-US} dataset to predict the number of natural deaths in each US state. As shown in Table \ref{table4}, all our three models again outperform several competitive baseline methods with significant margins. We also visualize and compare the prediction results with the true numbers at different weeks. It can be seen that our models can capture the main trend of natural death and predict the trend of the data more accurately. The details can be found in Appendix \ref{apd:Deaths}.



\begin{table}[t]
	\renewcommand{\arraystretch}{1.2}
	\centering
	\resizebox{.75\columnwidth}{!}{
	\begin{tabular}{c|c|c c c}
		\toprule[1.2pt]
		{Dataset} & Model & MAE & RMSE & MAPE  \\
		\midrule
		\multirow{6}{*}{DND-US} & AGCRN & 105.97 & 325.09 &  7.49\%  \\
		{} & DSTAGNN & 47.49 & 73.37 & 7.47\%  \\
		{} & STG-NCDE & 47.70 & 77.30 & 6.13\%  \\
		\cline{2-5}
		{} & \bf{A-ASTGCRN} & \bf{39.33} & \bf{62.86}* & \bf{5.36\%} \\
		{} & \bf{I-ASTGCRN} & \bf{38.79}* & \bf{66.99} & \bf{5.16\%}* \\
		{} & \bf{T-ASTGCRN} & \bf{40.60} & \bf{66.28} & \bf{5.43\%} \\
		\bottomrule[1.2pt]
	\end{tabular}
	}
    \vspace{-0.10in}
	\caption{Forecasting performance of several competitive methods on \emph{DND-US}}
	\label{table4}
 \vspace{-0.10in}
\end{table}

\subsection{Ablation and Parameter Study}
\begin{table}[t]
	\renewcommand{\arraystretch}{1.2}
	\centering
	\resizebox{.95\columnwidth}{!}{
	\begin{tabular}{c|c c c|c c c}
		\toprule[1.2pt]
		\multirow{2}{*}{Model} & \multicolumn{3}{c|}{PEMSD3} & \multicolumn{3}{c}{PEMSD4}\\
		\cline{2-7}
		{}& MAE & RMSE & MAPE & MAE & RMSE & MAPE  \\
		\midrule
		\bf{A-ANN} & 29.39 & 45.59 & 28.00\% & 35.99 & 52.52 & 26.36\%  \\
		\bf{I-ANN} & 20.63 & 33.93 & 20.37\% & 26.39 & 40.65 & 17.79\%  \\
		\bf{T-ANN} & 20.55 & 34.40 & 20.38\% & 26.02 & 40.04 & 17.74\%  \\
		\bf{STGCRN} & 17.69 & 30.53 & 16.76\% & 21.21 & 34.00 & 14.01\%  \\
            \bf{A-ASTGCRN(s)} & 17.41 & 29.49 & 16.00\% & 22.29 & 35.20 & 14.88\%  \\
            \bf{I-ASTGCRN(s)} & 17.38 & 29.37 & 16.63\% & 22.14 & 35.06 & 14.80\%  \\
            \bf{T-ASTGCRN(s)} & 17.39 & 29.52 & 15.85\% & 22.01 & 34.92 & 14.59\%  \\
		\hline
		\bf{A-ASTGCRN} & 15.06 & 26.71 & 13.83\%* & 19.30 & 30.92 & 12.91\%  \\
		\bf{I-ASTGCRN} & 15.06 & 26.40 & 13.91\% & 19.15* & 30.80* & 12.89\% \\
		\bf{T-ASTGCRN} & 14.90* & 26.01* & 14.17\% & 19.21 & 31.05 & 12.67\%*  \\
		\bottomrule[1.2pt]
	\end{tabular}
	}
    \vspace{-0.1in}
	\caption{Ablation experiments on \emph{PEMSD3} and \emph{PEMSD4}}
	\label{table5}
 \vspace{-0.10in}
\end{table}

{\noindent \bf Ablation Study.}
\noindent We refer to the model without an attention layer as \emph{STGCRN}, and the \emph{A-ASTGCRN}, \emph{I-ASTGCRN} and \emph{T-ASTGCRN} without the \emph{GCRN} layer as \emph{A-ANN}, \emph{I-ANN} and \emph{T-ANN}, respectively. Also, \emph{A-ASTGCRN(s)}, \emph{I-ASTGCRN(s)} and \emph{T-ASTGCRN(s)} are variant models that use static graphs for graph convolution. Table \ref{table5} shows the ablation experimental results on the \emph{PEMSD3} and \emph{PEMSD4} datasets. It shows that the performance of \emph{STGCRN} drops to that of a normal \emph{CNN} based approach, and the spatial modeling ability of the static graph is much less than that of the adaptive adjacency matrix. Moreover, the performance of the models with only the attention layer is extremely poor, especially for \emph{A-ANN}, which drops significantly and becomes similar as the traditional statistical methods. The attention module is crucial for capturing long-term temporal dependencies in traffic data, further enhancing the modeling of spatial-temporal dependencies. But, only focusing on long-term time dependence using the attention module and removing \emph{GRU} that uses adaptive graph convolution would damage the prediction performance. More detailed results are provided in the Appendix \ref{apd:horizon}.


\begin{table}[t]
	\renewcommand{\arraystretch}{1.2}
	\centering
	\resizebox{.95\columnwidth}{!}{
	\begin{tabular}{c|c|c c c c c c}
		\toprule[1.2pt]
		Dataset& {K}& MAE & RMSE & MAPE & Training & Inference & Memory \\
		\midrule
		\multirow{3}{*}{PEMSD3} & 1 & 15.24 & 26.46 & 15.10\% & \bf{88.94} & \bf{9.95} & \bf{6497} \\
		{} & \bf{2} & \bf{14.90} & \bf{26.01} & \bf{14.17}\% & 95.80 & 10.22 & 7555 \\
		{} & 3 & 15.33 & 27.04 & 13.92\% & 121.40 & 13.39 & 8535 \\
		\hline
		\multirow{3}{*}{PEMSD4} & 1 & 19.40 & 31.19 & 13.00\% & \bf{48.72} & \bf{5.24} & \bf{6355} \\
		{} & \bf{2} & \bf{19.21} & \bf{31.05} & \bf{12.67}\% & 54.80 & 5.62 & 7319 \\
		{} & 3 & 19.22 & 31.07 & 12.84\% & 66.43 & 7.12 & 8137 \\
		\bottomrule[1.2pt]
	\end{tabular}
	}
        \vspace{-0.10in}
	\caption{Effect of convolution kernel number $K$ on \emph{T-ASTGCRN}.}
	\label{tablep1}
  \vspace{-0.10in}
\end{table}

\begin{figure}[t]
	\centering
	\begin{subfigure}{0.48\linewidth}
		\centering
		\includegraphics[width=1\linewidth]{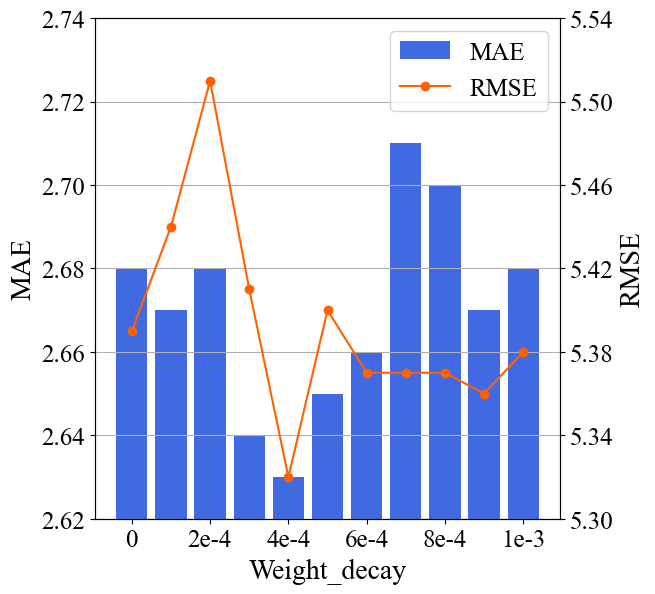}
		\caption{Effects of varied weight decay value}
	\end{subfigure}
	\begin{subfigure}{0.48\linewidth}
		\centering
		\includegraphics[width=1\linewidth]{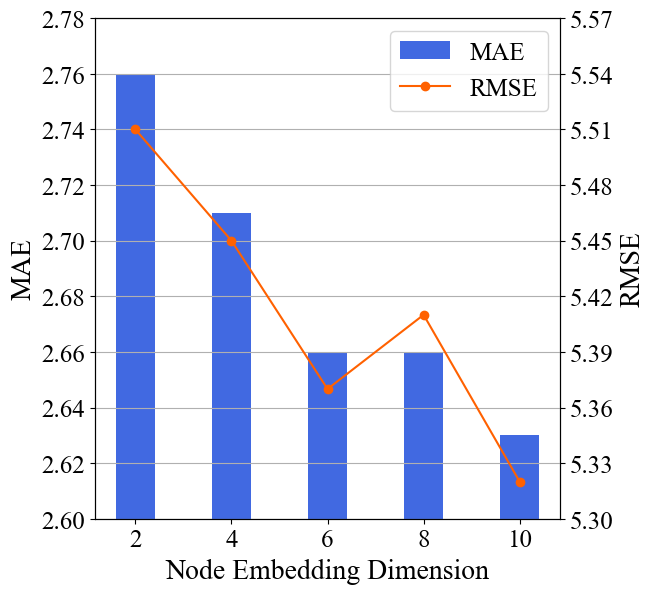}
		\caption{Effects of varied node embedding dimension ($D_{e}$)}
	\end{subfigure}
    \vspace{-0.10in}
	\caption{Effects of hyperparameter tuning on \emph{T-ASTGCRN} in \emph{PEMSD7(M)}}
	\label{fig3}
 \vspace{-0.10in}
\end{figure}

\smallskip
{\noindent \bf Parameter Study.}
\noindent To investigate the effects of hyperparameters on the prediction results, we conduct a series of experiments on the main hyperparameters. Figure \ref{tablep1} shows the prediction performance and training cost for varying the number of convolution kernels $K$ on the PEMSD3 and PEMSD4 datasets. From the experimental results, we can see that with $K=1$, the graph convolution is simplified to a unit matrix-based implementation, which does not enable effective information transfer between nodes. A larger convolution depth does not improve the prediction performance, but instead incurs longer training time and memory cost. Therefore, for our model and dataset, we set $K$ to $2$. Meanwhile, Figure \ref{fig3} shows the \emph{MAE} and \emph{RMSE} values of \emph{T-ASTGCRN} in the \emph{PEMSD7(M)} dataset when varying the weight decay and node embedding dimension $D_{e}$. It can be seen that increasing the weight decay and node embedding dimension appropriately can improve the prediction performance of \emph{T-ASTGCRN}. However, the weight decay should not be too high, as otherwise the performance of the model could be significantly reduced. When the weight decay is $0.0004$ and $D_{e}=10$, the two performance metrics reach their lowest values.

\section{Conclusion}
\noindent In this paper, we design an attention-based spatial-temporal graph convolutional recurrent network framework for traffic prediction. We instantiate the framework with three attention modules based on Multi-head self-attention, Transformer and Informer, all of which, in particular the Transformer-based module, can well capture long-term temporal dependence and incorporate with the spatial and short-term temporal features by the \emph{GCRN} module. Extensive experiments confirm the effectiveness of all our three models in improving the prediction performance. We believe that the design ideas of Transformer and Informer can bring new research thrusts in the field of traffic forecasting.

\bibliographystyle{named}
\bibliography{ijcai22bib}

\clearpage
\appendix
\renewcommand{\appendixname}{Appendix~\Alph{section}}

\section{Baselines Information}
\label{apd:baseline}
\noindent We compare our models with the following baseline models:
\begin{itemize}
\item \textbf{\emph{HA}}: Historical Average models traffic flow as a periodic process and uses the average of historical traffic flow (eg, the same time in previous weeks) to predict future traffic flow.

\item \textbf{\emph{ARIMA}}: Auto-Regressive Integrated Moving Average method, which is a widely used model for time series forecasting \cite{williams2003modeling}.

\item \textbf{\emph{VAR}}: Vector Auto-Regression is a statistical model that captures the relationship of multiple variables over time \cite{zivot2006vector}.

\item \textbf{\emph{SVR}}: Support Vector Regression is a traditional time series forecasting model that uses a linear support vector machine for regression tasks \cite{drucker1996support}.

\item \textbf{\emph{FC-LSTM}}: \emph{LSTM} network with fully connected hidden units, which is a network model that can effectively capture time dependencies \cite{sutskever2014sequence}.

\item \textbf{\emph{DCRNN}}: Diffusion Convolutional Recurrent Neural Network, which captures spatial and temporal dependencies using diffuse graph convolution and encoder-decoder network architecture, respectively \cite{li2018dcrnn_traffic}.

\item \textbf{\emph{STGCN}}: Spatial-Temporal Graph Convolutional Network, which combines graph convolution and 1D convolution to capture spatial-temporal correlations \cite{yu2018spatio}.

\item \textbf{\emph{Graph WaveNet}}: Graph WaveNet introduces an adaptive adjacency matrix and combines diffuse graph convolution with 1D convolution \cite{wu2019graph}.

\item \textbf{\emph{ASTGCN(r)}}: Attention Based Spatial-Temporal Graph Convolutional Networks, which fuses spatial attention and temporal attention mechanisms with spatial-temporal convolution to capture dynamic spatial-temporal features. we use the latest components to ensure the fairness of the comparison \cite{guo2019attention}.

\item \textbf{\emph{MSTGCN}}: Multi-Component Spatial-Temporal Graph Convolution Networks, which are \emph{ASTGCN(r)} that discard the spatial-temporal attention mechanism.

\item \textbf{\emph{LSGCN}}: Long Short-term Graph Convolutional Networks, which proposes a new graph attention network and integrates it with graph convolution into a spatial gated block \cite{huang2020lsgcn}.

\item \textbf{\emph{STSGCN}}: Spatial-Temporal Synchronous Graph Convolutional Networks, which enables the model to efficiently extract localized spatial-temporal correlations through a well-designed local spatial-temporal subgraph module \cite{song2020spatial}.

\item \textbf{\emph{AGCRN}}: Adaptive Graph Convolutional Recurrent Network, which augments traditional graph convolution with adaptive graph generation and node adaptive parameter learning, and is integrated into a recurrent neural network to capture more complex spatial-temporal correlations \cite{bai2020adaptive}.

\item \textbf{\emph{STFGNN}}: Spatial-Temporal Fusion Graph Neural Networks, which designs a new spatial-temporal fusion graph module and assembles it in parallel with 1D convolution module to capture both local and global spatial-temporal dependencies \cite{li2021spatial}.

\item \textbf{\emph{ASTGNN}}: Attention based Spatial-Temporal Graph Neural Network, which proposes a new method for spatial-temporal modeling of traffic data dynamics, taking into account the periodicity and spatial heterogeneity of traffic data \cite{guo2021learning}.

\item \textbf{\emph{STGODE}}: Spatial-Temporal Graph Ordinary Differential Equation Networks, which captures spatial-temporal dynamics through a tensor-based ordinary differential equation (\emph{ODE}) \cite{fang2021spatial}.

\item \textbf{\emph{Z-GCNETs}}: Time Zigzags at Graph Convolutional Networks, which introduces the concept of Zigzag persistence to time-aware graph convolutional networks \cite{chen2021z}.

\item \textbf{\emph{STG-NCDE}}: Spatial-Temporal Graph Neural Controlled Differential Equation, which designs two NCDEs for temporal processing and spatial processing and integrates them into a single framework \cite{choi2022graph}.

\item \textbf{\emph{DSTAGNN}}: Dynamic Spatial-Temporal Aware Graph Neural Network, which proposes a new dynamic spatial-temporal awareness graph to replace the predefined static graph used by traditional graph convolution \cite{lan2022dstagnn}.

\end{itemize}

\section{Best Hyperparameters}
\label{apd:hyperparam}
\noindent Following are the hyperparameter configurations for our three attention-based spatial-temporal graph convolutional recurrent neural networks (\emph{i.e.}, \emph{A-ASTGCRN}, \emph{T-ASTGCRN}, \emph{I-STGCRN}) to achieve optimal performance on each dataset:

\begin{itemize}
    \item \textbf{\emph{A-ASTGCRN}}: In the \emph{PEMSD3} dataset, the dataset batch size is $64$, the weight decay coefficient is $0.0009$ and the node embedding dimension is $10$; in the \emph{PEMSD4} dataset, the dataset batch size is $64$, the weight decay coefficient is $0.0004$ and the node embedding dimension is $4$; in the \emph{PEMSD7} dataset, the dataset batch size is $16$, the weight decay coefficient is $0.0003$ and node embedding dimension of $10$; in the \emph{PEMSD8} dataset, the dataset batch size is $64$, the weight decay factor is $0.0001$ and the node embedding dimension is $2$; in the \emph{PEMSD7(M)} dataset, the dataset batch size is $64$, the weight decay factor is $0.0008$ and the node embedding dimension is $10$, in the \emph{DND-US} dataset, the dataset batch size is $8$, the weight decay factor is $0.0001$ and the node embedding dimension is $10$. 
    
    \item \textbf{\emph{T-ASTGCRN}}: In the \emph{PEMSD3} dataset, the dataset batch size is $64$, the weight decay coefficient is $0.0009$ and the node embedding dimension is $10$; in the \emph{PEMSD4} dataset, the dataset batch size is $64$, the weight decay coefficient is $0.001$ and the node embedding dimension is $10$; in the \emph{PEMSD7} dataset, the dataset batch size is $16$, the weight decay coefficient is $0.0003$ and node embedding dimension of $10$; in the \emph{PEMSD8} dataset, the dataset batch size is $64$, the weight decay factor is $0.0004$ and the node embedding dimension is $2$; in the \emph{PEMSD7(M)} dataset, the dataset batch size is $64$, the weight decay factor is $0.0004$ and the node embedding dimension is $10$, in the \emph{DND-US} dataset, the dataset batch size is $8$, the weight decay factor is $0.0001$ and the node embedding dimension is $10$.
    
    \item \textbf{\emph{I-ASTGCRN}}: In the \emph{PEMSD3} dataset, the dataset batch size is $64$, the weight decay coefficient is $0.0005$ and the node embedding dimension is $8$; in the \emph{PEMSD4} dataset, the dataset batch size is $64$, the weight decay coefficient is $0.0004$ and the node embedding dimension is $4$; in the \emph{PEMSD7} dataset, the dataset batch size is $16$, the weight decay coefficient is $0.0005$ and node embedding dimension of $10$; in the \emph{PEMSD8} dataset, the dataset batch size is $64$, the weight decay factor is $0.001$ and the node embedding dimension is $2$. in the \emph{PEMSD7(M)} dataset, the dataset batch size is $64$, the weight decay factor is $0.0004$ and the node embedding dimension is $10$, in the \emph{DND-US} dataset, the dataset batch size is $8$, the weight decay factor is $0.0001$ and the node embedding dimension is $10$. 
\end{itemize}

The learning rate on all datasets is $0.003$, and the number of convolution kernels $K=2$. The \emph{GCRN} number of layers is $2$, where the number of hidden units per layer in the traffic datasets is $64$, while that in \emph{DND-US} is $32$.

\section{Prediction Metrics}
\label{apd:metrics}
\noindent We use three common metrics to evaluate the performance of all models: Mean Absolute Error (\emph{MAE}), Root Mean Square Error (\emph{RMSE}), Mean Absolute Percent Error (\emph{MAPE}). Their formal definitions are as follows:
\begin{equation}
\begin{aligned}
&\operatorname{MAE}(\hat{Y}, Y)=\frac{1}{T} \sum_{i=1}^{T}\left|\hat{y}_{i}-y_{i}\right| \\
&\operatorname{RMSE}(\hat{Y}, Y)=\sqrt{\frac{1}{T} \sum_{i=1}^{T}\left(\hat{y}_{i}-y_{i}\right)^{2}} \\
&\operatorname{MAPE}(\hat{Y}, Y)=\frac{100\%}{T} \sum_{i=1}^{T}\left|\frac{\hat{y}_{i}-y_{i}}{\hat{y}_{i}}\right|
\end{aligned}
\end{equation}
where $\hat{Y}=\hat{y}_{1}, \hat{y}_{2}, \dots, \hat{y}_{T}$ is the real traffic data, $Y={y}_{1}, {y}_{2}, \dots, {y}_{T}$ is the predicted data, and $T$ is the predicted time step. In our experiments, $T=12$.

\begin{table*}[t]
	\renewcommand{\arraystretch}{1.2}
	\centering
	\resizebox{.75\linewidth}{!}{
	\begin{tabular}{c|c c c|c c c |c c c}
		\toprule[1.2pt]
		\multirow{2}{*}{Model} & \multicolumn{3}{c|}{3 week} & \multicolumn{3}{c|}{6 week} & \multicolumn{3}{c}{12 week} \\
		\cline{2-10}
		{}& MAE & RMSE & MAPE & MAE & RMSE & MAPE & MAE & RMSE & MAPE   \\
		\midrule
		AGCRN & 104.57 & 322.01 & 7.37\% & 101.26 & 311.41 & 7.32\% & 115.46 & 347.93 & 7.99\%  \\
		DSTAGNN & 40.83 & 65.58 & 6.58\% & 44.42 & 65.88 & 7.31\% & 59.32 & 89.61 & 8.72\%  \\
		STG-NCDE & 46.81 & 75.06 & 6.14\% & 45.56 & 72.18 & 5.93\% & 50.02 & 83.80 & 6.23\%  \\
		\midrule
		\bf{A-ASTGCRN} & \bf{37.51} & \bf{60.77} & \bf{5.21\%} & \bf{38.17} & \bf{58.62}* & \bf{5.26\%} & \bf{43.70}* & \bf{72.16}* & \bf{5.74\%}* \\
		\bf{I-ASTGCRN} & \bf{35.29} & \bf{59.99} & \bf{4.91\%}* & \bf{35.67}* & \bf{61.93} & \bf{4.87\%}* & \bf{46.23} & \bf{80.44} & \bf{5.78\%} \\
		\bf{T-ASTGCRN} & \bf{34.20}* & \bf{55.48}* & \bf{4.93\%} & \bf{40.49} & \bf{65.20} & \bf{5.40\%} & \bf{48.34} & \bf{78.87} & \bf{6.09\%} \\
		\bottomrule[1.2pt]
	\end{tabular}
	}
	\caption{Forecasting results on DND-US.}
	\label{tableDND}
\end{table*}


\section{Case Study on DND-US}
\label{apd:Deaths}
\noindent To test the generalizability of our proposed model for different spatial-temporal learning tasks, we conduct an additional experiment on the US natural death dataset DND-US and analyzed the results in detail. We use weekly natural death data from 01/04/2014 to 12/28/2019 for $53$ states or autonomous states in the United States and divide them into training, testing, and prediction sets in the ratio of $6$:$2$:$2$. Both our model and baseline methods use $12$ consecutive time steps ($12$ weeks) of data to predict the next $12$ consecutive time steps of data. Table \ref{tableDND} shows the comparative performance of different methods for predicting week $3$, week $6$, and week $12$, where all our three models outperform the other three competitive baseline methods.

\begin{figure*}[t]
	\centering
	\begin{subfigure}{0.45\linewidth}
		\centering
		\includegraphics[width=1\linewidth]{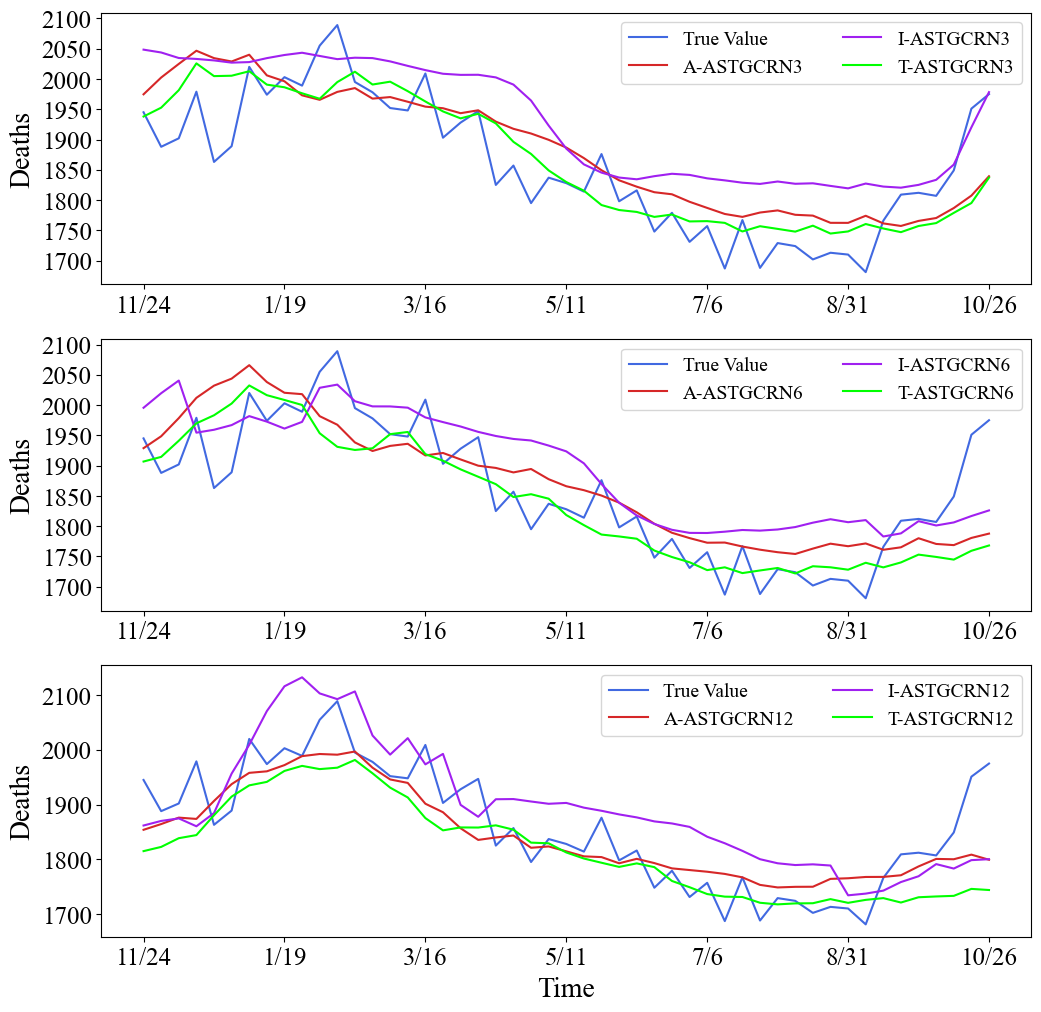}
		\caption{Forecast results for week $3$, week $6$, and week $12$}
	\end{subfigure}
	\begin{subfigure}{0.45\linewidth}
		\centering
		\includegraphics[width=1\linewidth]{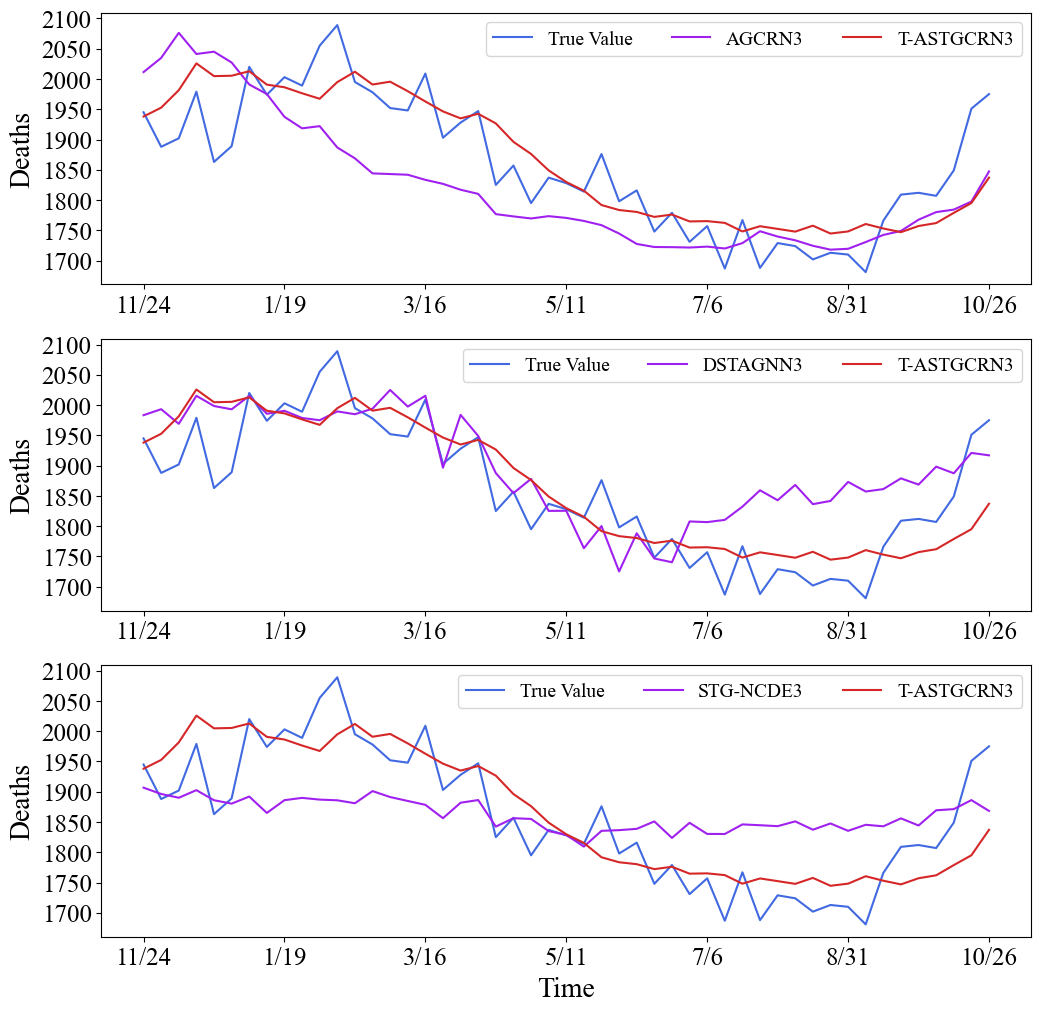}
		\caption{\scriptsize Forecast results for week $3$ of \emph{T-ASTGCRN} and baseline methods}
	\end{subfigure}
	
	\begin{subfigure}{0.45\linewidth}
		\centering
		\includegraphics[width=1\linewidth]{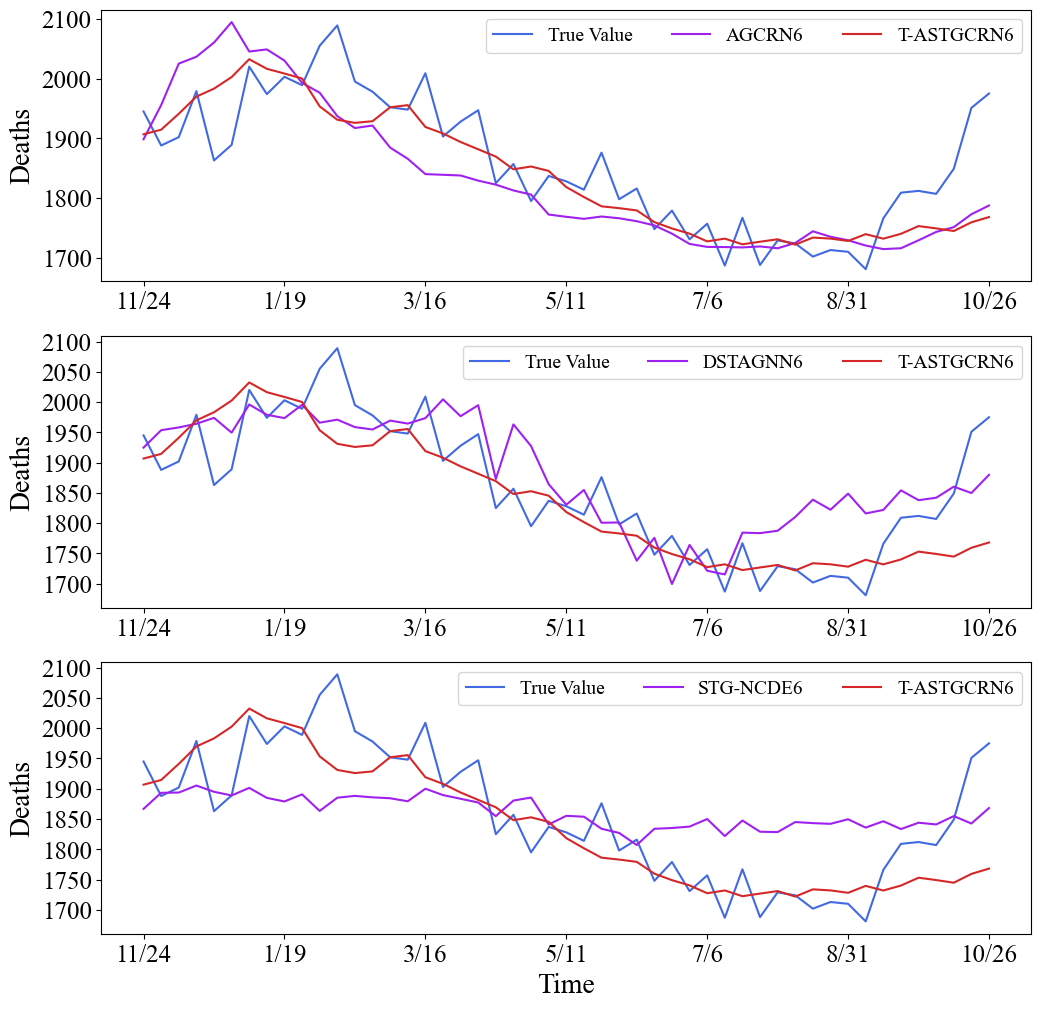}
		\caption{Forecast results for week $6$ of \emph{T-ASTGCRN} and baseline methods}
	\end{subfigure}
	\begin{subfigure}{0.45\linewidth}
		\centering
		\includegraphics[width=1\linewidth]{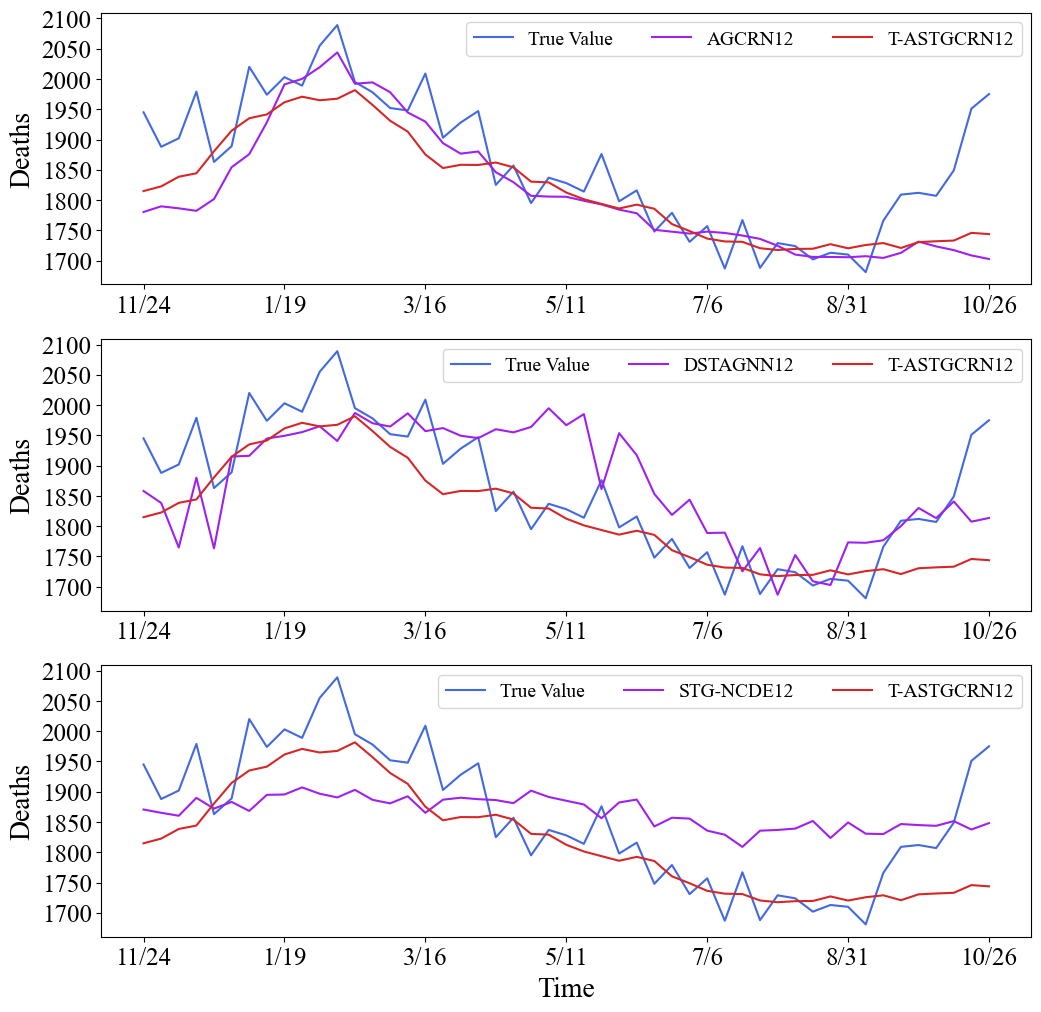}
		\caption{Forecast results for week $12$ of \emph{T-ASTGCRN} and baseline methods}
	\end{subfigure}
	\caption{Visualization of forecast results for node $13$ (Illinois) on \emph{DND-US}}
	\label{figDND1}
\end{figure*}

Accurate prediction of mortality trends and numbers helps governments to evaluate their impact in advance and design effective public health policies. Taking node $13$ (Illinois) as an example, Figure 4(a) illustrates the prediction results of our three models for week $3$ (with suffix $3$), week $6$ (with suffix $6$), and week $12$ (with suffix $12$). As shown in the figure, the trends predicted by our models well match the real numbers. To observe the performance difference between our model and baseline methods more clearly, we visualize and compare our representative method \emph{T-ASTGCRN} with several baseline methods. Figures \ref{figDND1}(b), \ref{figDND1}(c), and \ref{figDND1}(d) show their prediction results for week $3$, week $6$, and week $12$, respectively. It can be seen that our \emph{T-ASTGCRN} can capture the main trend of natural death and predict the trend of the data more accurately. Compared with \emph{T-ASTGCRN}, the prediction results of the baseline methods are much different from the true values and have significant delays in data changes.

\section{Ablation Experiments}
\label{apd:horizon}
\begin{figure*}[b]
    \centering
         \begin{subfigure}{0.30\linewidth}
        \centering
        \includegraphics[width=1\linewidth]{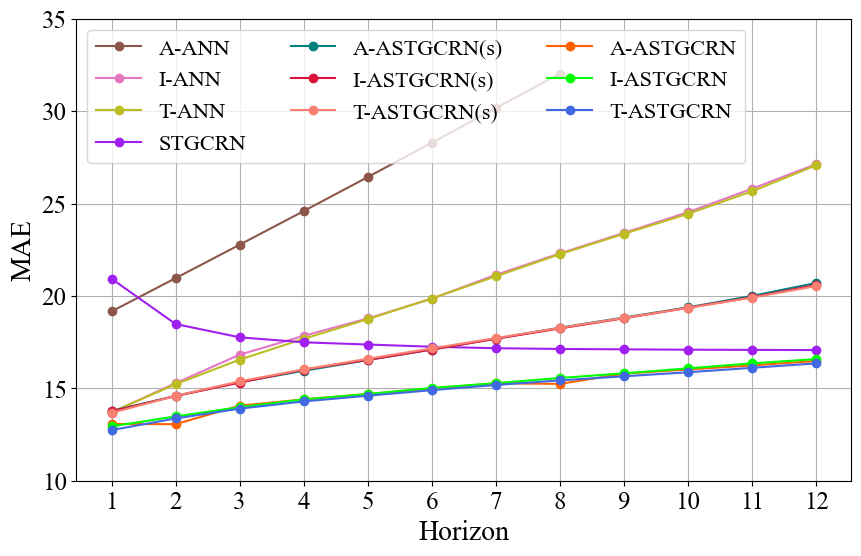}
        \caption{\emph{MAE} on \emph{PEMSD3}}
    \end{subfigure}
    \begin{subfigure}{0.30\linewidth}
        \centering
        \includegraphics[width=1\linewidth]{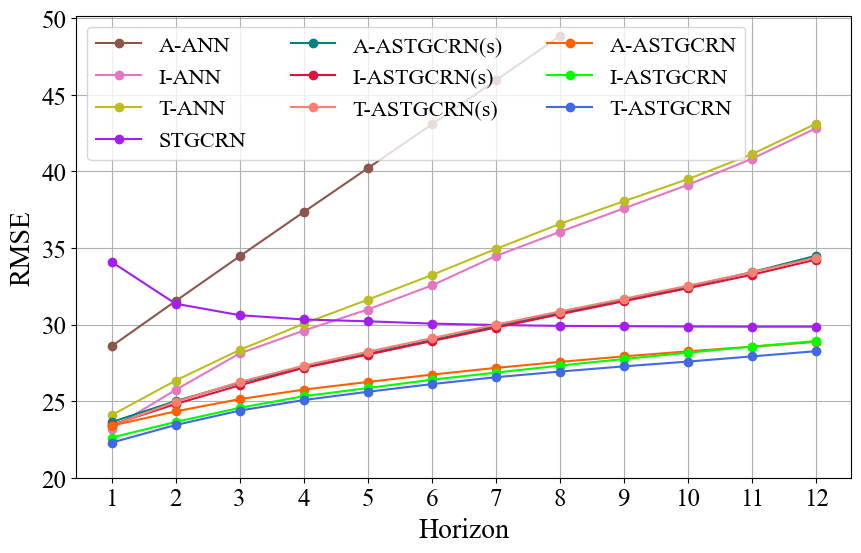}
        \caption{\emph{RMSE} on \emph{PEMSD3}}
    \end{subfigure}
    \begin{subfigure}{0.30\linewidth}
        \centering
        \includegraphics[width=1\linewidth]{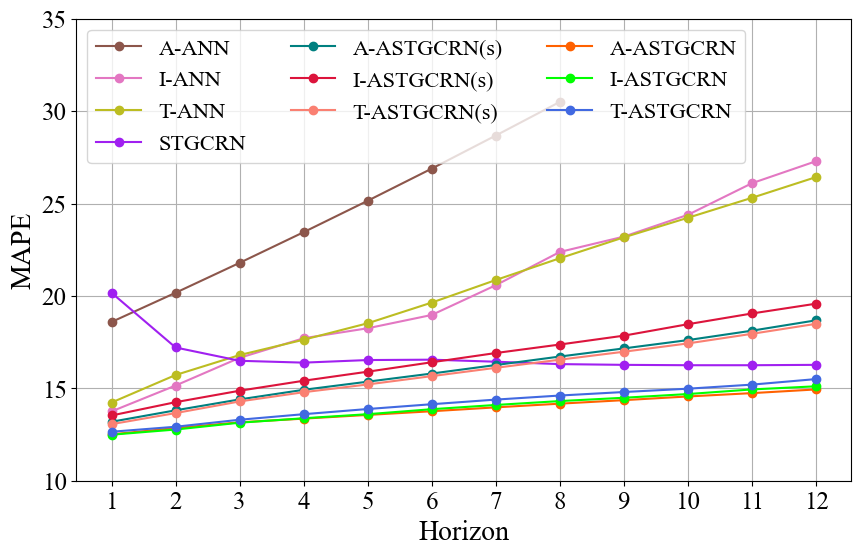}
        \caption{\emph{MAPE} on \emph{PEMSD3}}
    \end{subfigure}

        \begin{subfigure}{0.30\linewidth}
        \centering
        \includegraphics[width=1\linewidth]{figures/PEMSD3_MAE2.png}
        \caption{\emph{MAE} on \emph{PEMSD4}}
    \end{subfigure}
    \begin{subfigure}{0.30\linewidth}
        \centering
        \includegraphics[width=1\linewidth]{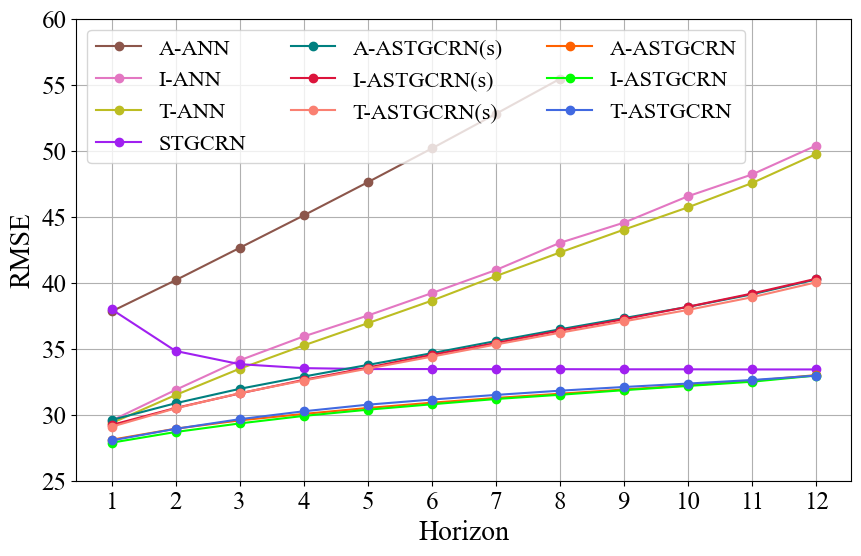}
        \caption{\emph{RMSE} on \emph{PEMSD4}}
    \end{subfigure}
    \begin{subfigure}{0.30\linewidth}
        \centering
        \includegraphics[width=1\linewidth]{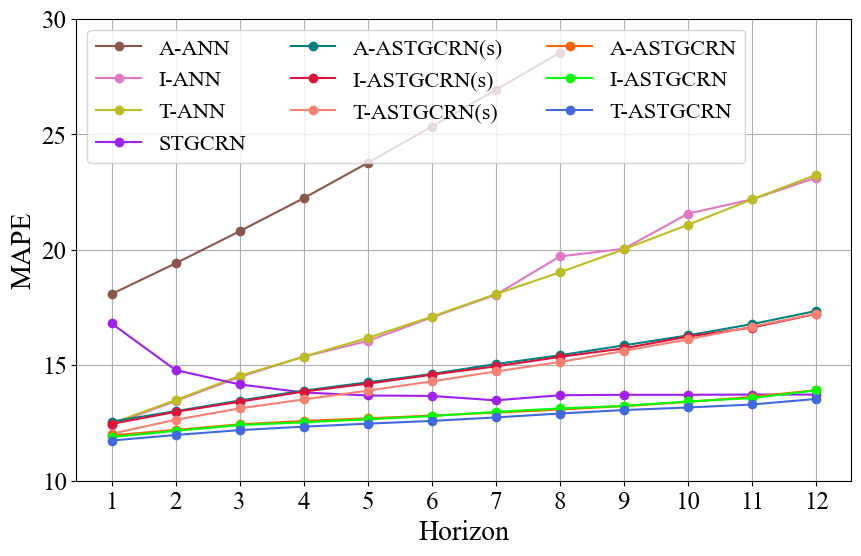}
        \caption{\emph{MAPE} on \emph{PEMSD4}}
    \end{subfigure}
    \caption{Prediction performance at each horizon}
    \label{fig4}
\end{figure*}

\noindent We plot the detailed values of \emph{MAE} of different horizons for our methods on the PEMSD3 and PEMSD4 in Figure \ref{fig4}. It shows that the \emph{MAE} values of \emph{STGCRN} become closer to the three models as the predicted horizon increases. The autoregressive feature of the \emph{GRU} model allows more spatial-temporal information to be pooled in the later time horizons, so that long-term prediction appears to be better than short-term prediction. But the performance of \emph{STGCRN} lags behind the three attention-based models at all time horizons. 

\section{Hyperparameters Analysis}
\begin{figure*}[t]
	\centering
	\begin{subfigure}{0.19\linewidth}
		\centering
		\includegraphics[width=1\linewidth]{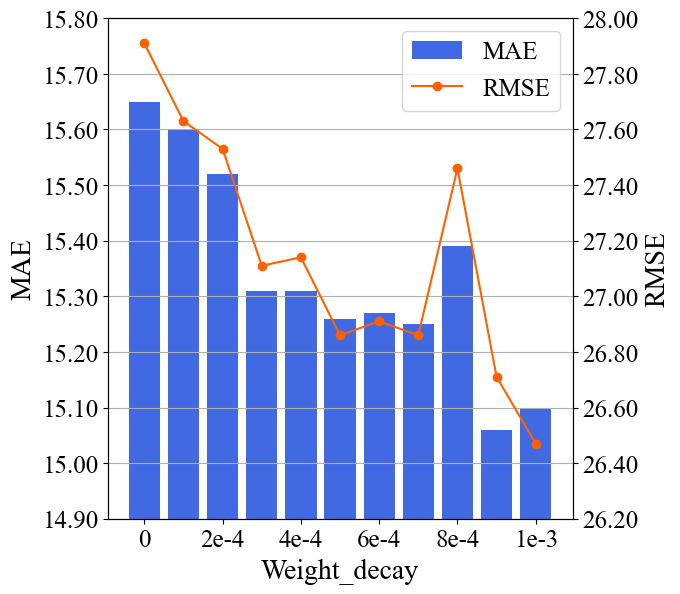}
		\caption{Effects of weight\_decay on \emph{PEMSD3}}
	\end{subfigure}
	\begin{subfigure}{0.19\linewidth}
		\centering
		\includegraphics[width=1\linewidth]{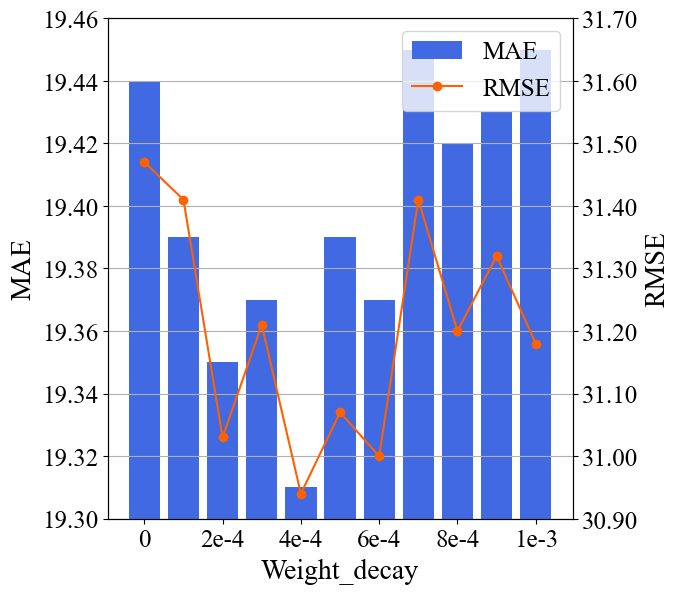}
		\caption{Effects of weight\_decay on \emph{PEMSD4}}
	\end{subfigure}
	\begin{subfigure}{0.19\linewidth}
		\centering
		\includegraphics[width=1\linewidth]{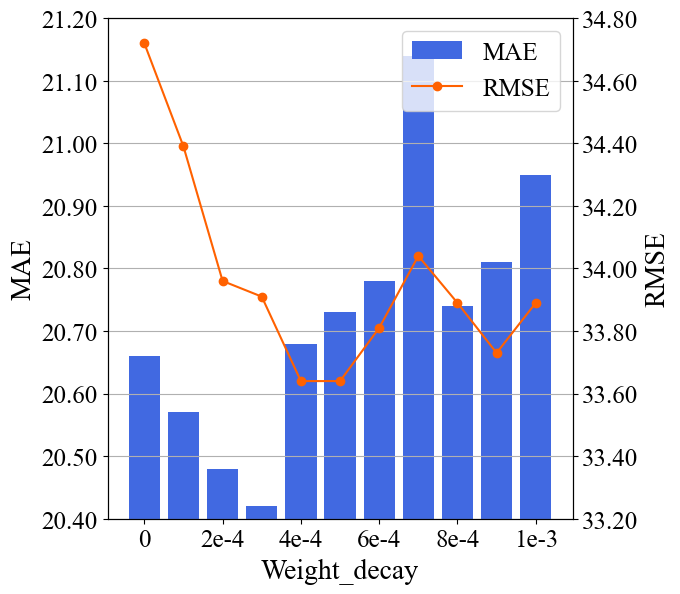}
		\caption{Effects of weight\_decay on \emph{PEMSD7}}
	\end{subfigure}
	\begin{subfigure}{0.19\linewidth}
		\centering
		\includegraphics[width=1\linewidth]{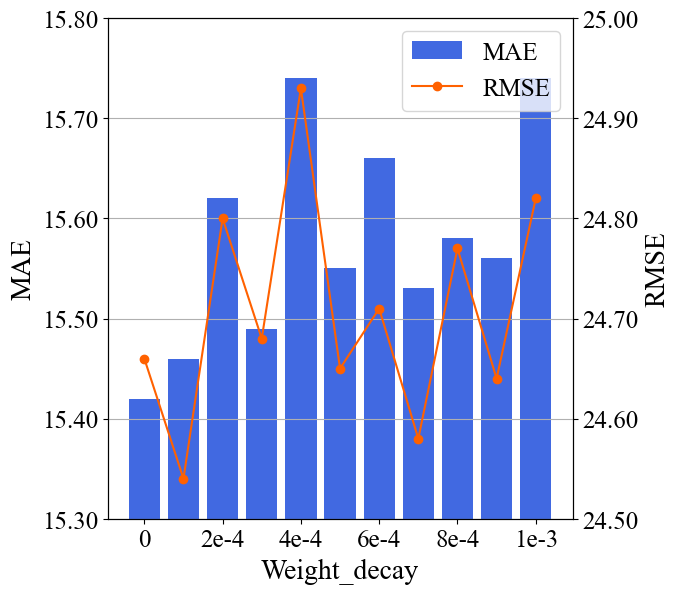}
		\caption{Effects of weight\_decay on \emph{PEMSD8}}
	\end{subfigure}
	\begin{subfigure}{0.19\linewidth}
		\centering
		\includegraphics[width=1\linewidth]{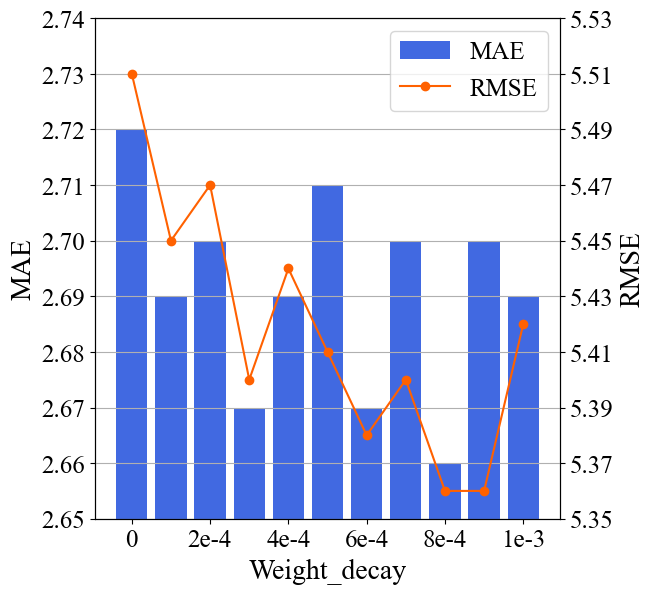}
		\caption{\scriptsize Effects of weight\_decay on \emph{PEMSD7(M)}}
	\end{subfigure}

	\begin{subfigure}{0.19\linewidth}
		\centering
		\includegraphics[width=1\linewidth]{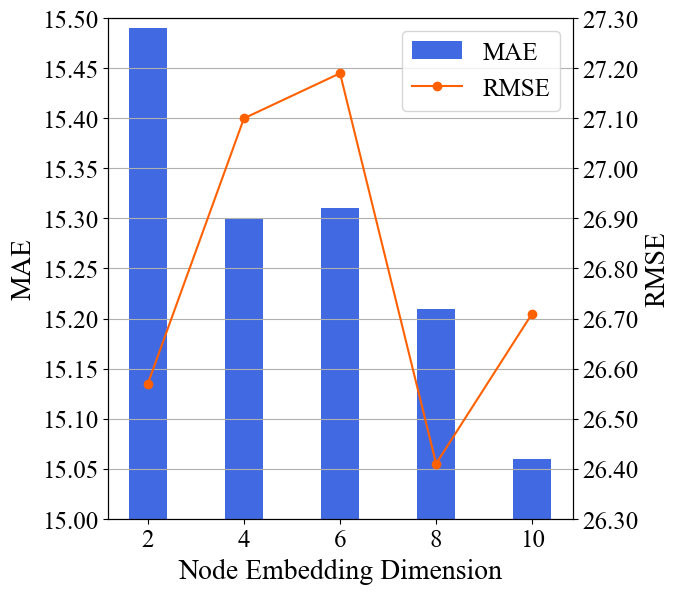}
		\caption{Effects of node embedding dimension ($D_{e}$) on \emph{PEMSD3}}
	\end{subfigure}
	\begin{subfigure}{0.19\linewidth}
		\centering
		\includegraphics[width=1\linewidth]{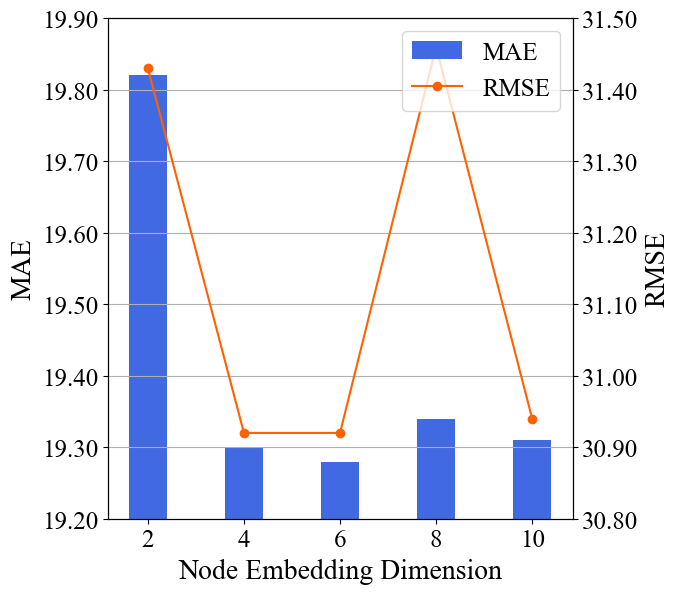}
		\caption{Effects of node embedding dimension ($D_{e}$) on \emph{PEMSD4}}
	\end{subfigure}
	\begin{subfigure}{0.19\linewidth}
		\centering
		\includegraphics[width=1\linewidth]{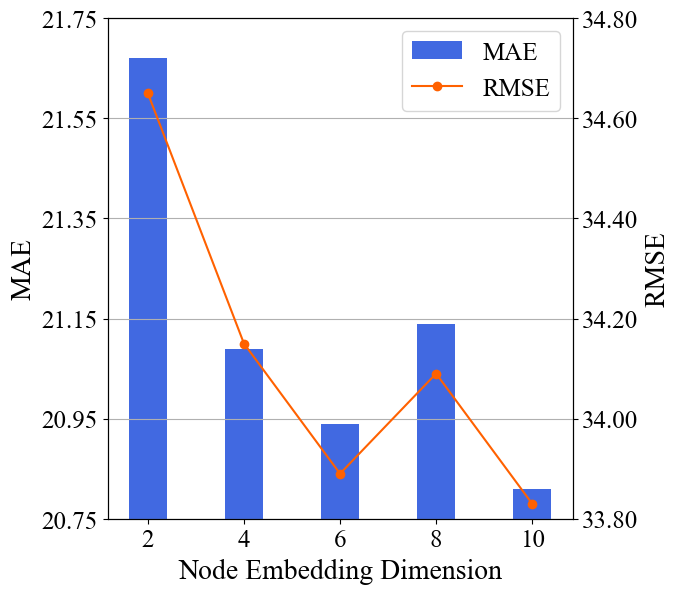}
		\caption{Effects of node embedding dimension ($D_{e}$) on \emph{PEMSD7}}
	\end{subfigure}
	\begin{subfigure}{0.19\linewidth}
		\centering
		\includegraphics[width=1\linewidth]{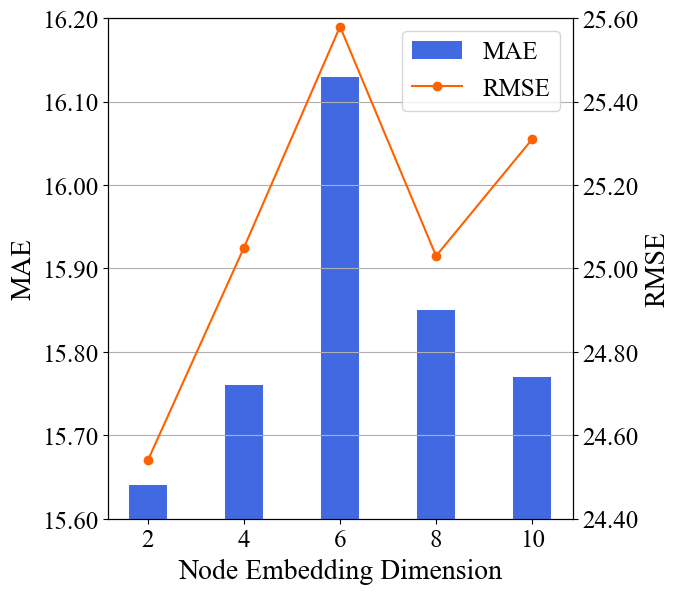}
		\caption{Effects of node embedding dimension ($D_{e}$) on \emph{PEMSD8}}
	\end{subfigure}
	\begin{subfigure}{0.19\linewidth}
		\centering
		\includegraphics[width=1\linewidth]{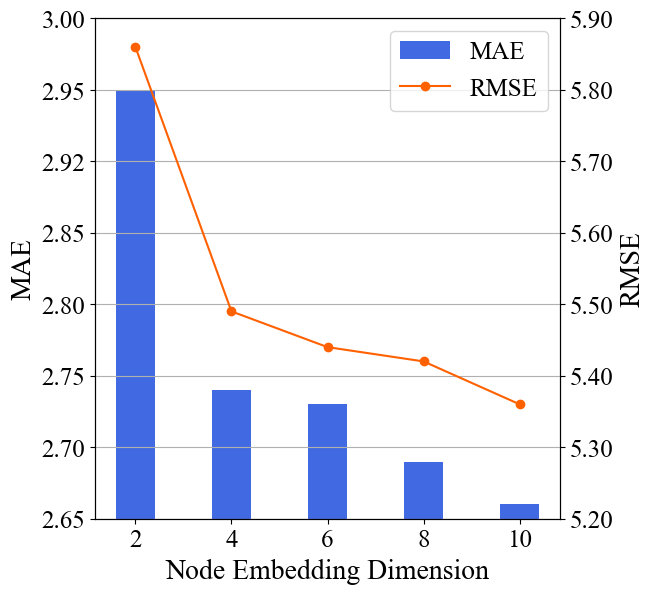}
		\caption{Effects of node embedding dimension ($D_{e}$) on \emph{PEMSD7(M)}}
	\end{subfigure}
	\caption{Hyperparameter experiments of \emph{A-ASTGCRN}}
	\label{fig5}
\end{figure*}

\begin{figure*}[t]
	\centering
	\begin{subfigure}{0.19\linewidth}
		\centering
		\includegraphics[width=1\linewidth]{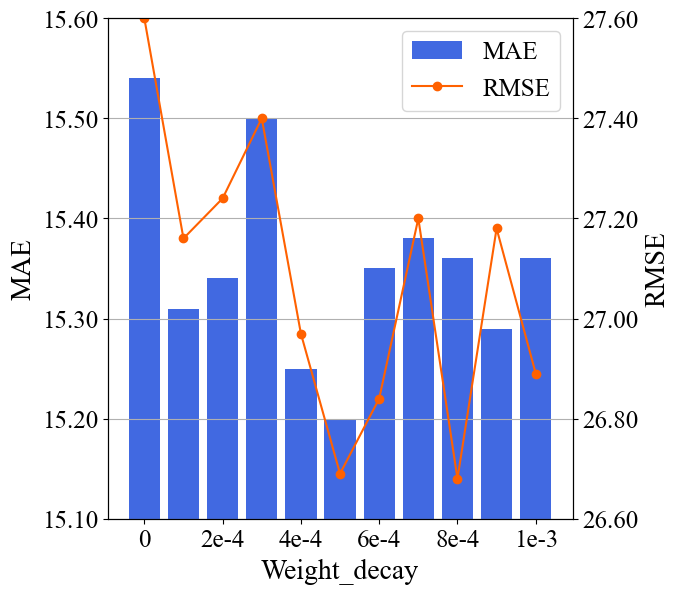}
		\caption{Effects of weight\_decay on \emph{PEMSD3}}
	\end{subfigure}
	\begin{subfigure}{0.19\linewidth}
		\centering
		\includegraphics[width=1\linewidth]{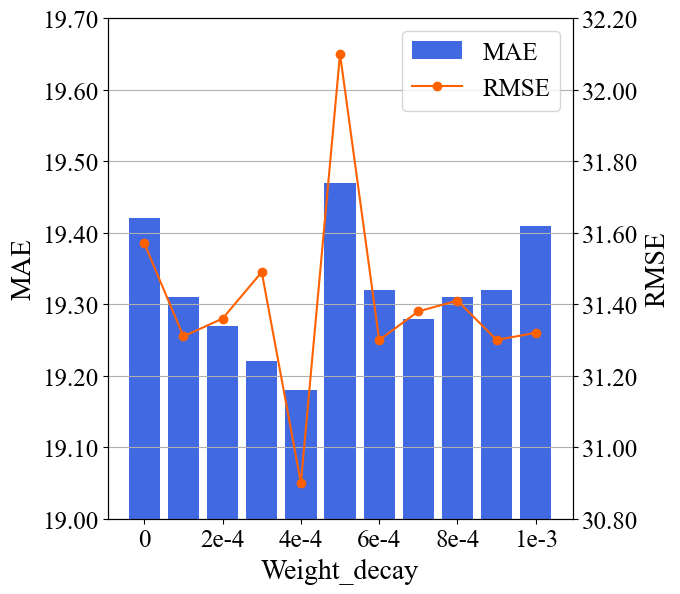}
		\caption{Effects of weight\_decay on \emph{PEMSD4}}
	\end{subfigure}
	\begin{subfigure}{0.19\linewidth}
		\centering
		\includegraphics[width=1\linewidth]{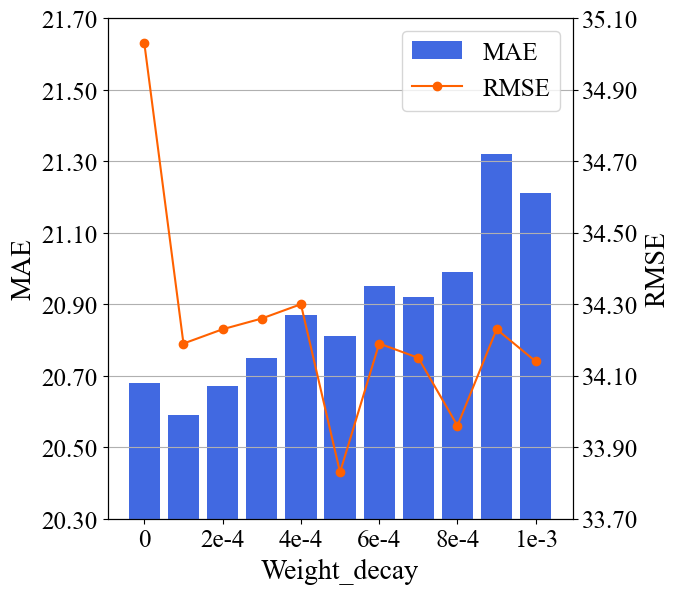}
		\caption{Effects of weight\_decay on \emph{PEMSD7}}
	\end{subfigure}
	\begin{subfigure}{0.19\linewidth}
		\centering
		\includegraphics[width=1\linewidth]{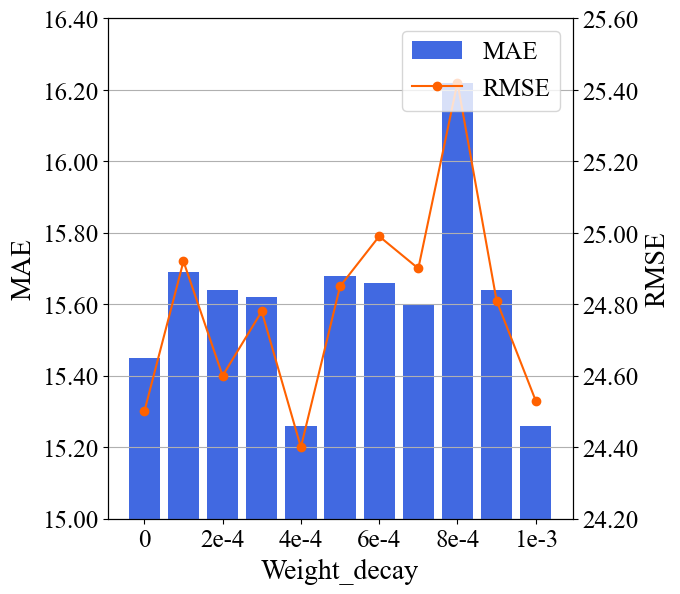}
		\caption{Effects of weight\_decay on \emph{PEMSD8}}
	\end{subfigure}
	\begin{subfigure}{0.19\linewidth}
		\centering
		\includegraphics[width=1\linewidth]{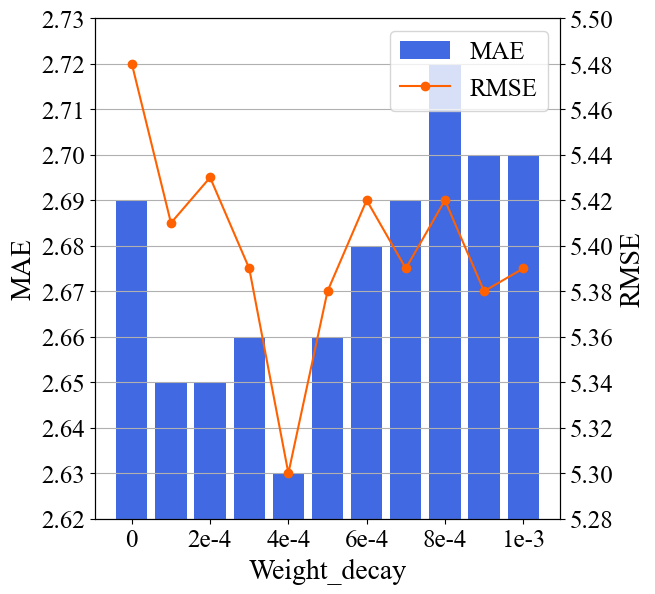}
		\caption{\scriptsize Effects of weight\_decay on \emph{PEMSD7(M)}}
	\end{subfigure}

	\begin{subfigure}{0.19\linewidth}
		\centering
		\includegraphics[width=1\linewidth]{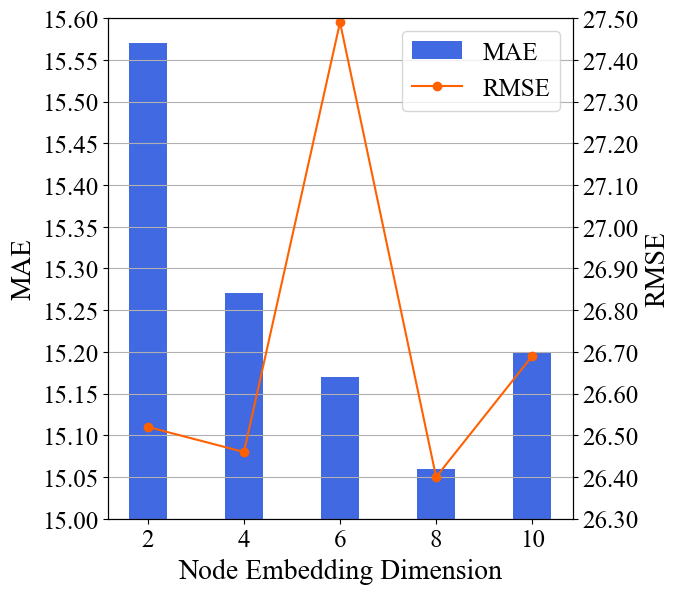}
		\caption{Effects of node embedding dimension ($D_{e}$) on \emph{PEMSD3}}
	\end{subfigure}
	\begin{subfigure}{0.19\linewidth}
		\centering
		\includegraphics[width=1\linewidth]{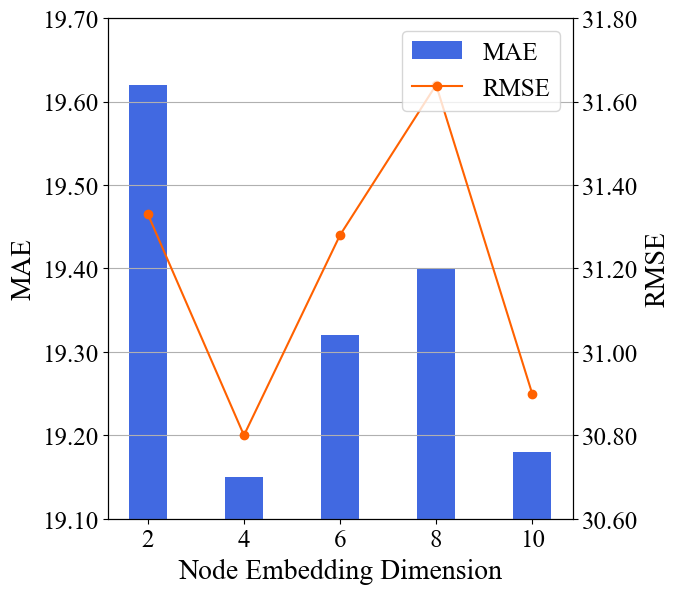}
		\caption{Effects of node embedding dimension ($D_{e}$) on \emph{PEMSD4}}
	\end{subfigure}
	\begin{subfigure}{0.19\linewidth}
		\centering
		\includegraphics[width=1\linewidth]{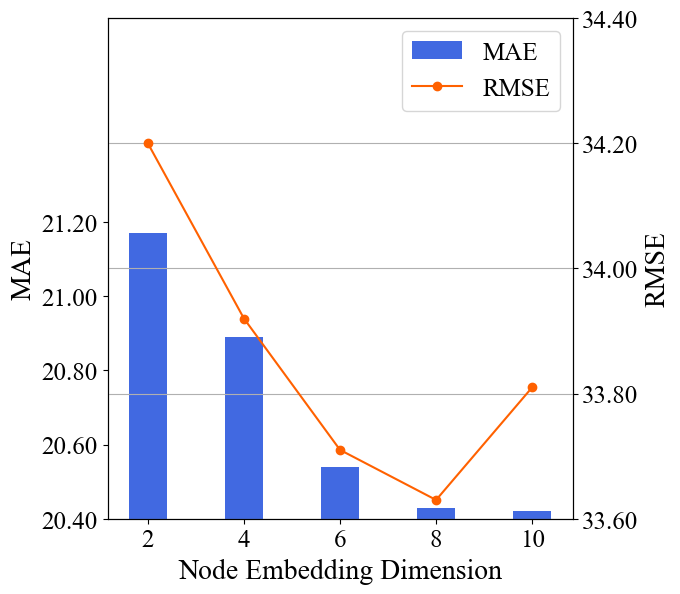}
		\caption{Effects of node embedding dimension ($D_{e}$) on \emph{PEMSD7}}
	\end{subfigure}
	\begin{subfigure}{0.19\linewidth}
		\centering
		\includegraphics[width=1\linewidth]{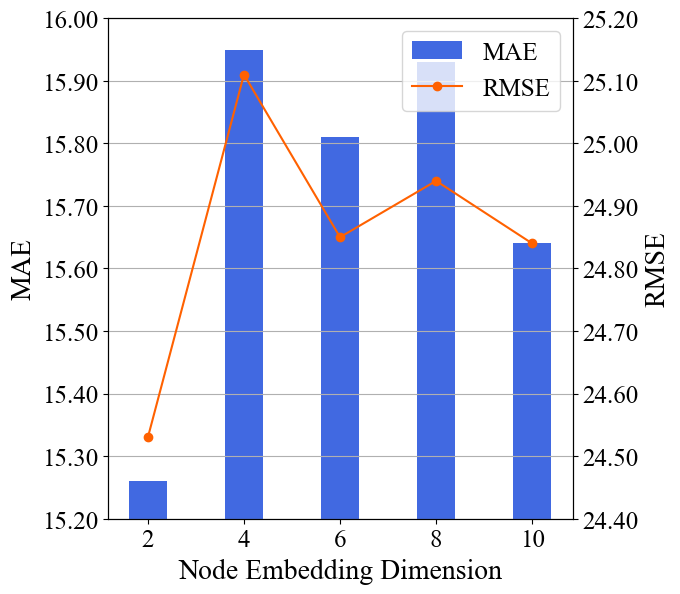}
		\caption{Effects of node embedding dimension ($D_{e}$) on \emph{PEMSD8}}
	\end{subfigure}
	\begin{subfigure}{0.19\linewidth}
		\centering
		\includegraphics[width=1\linewidth]{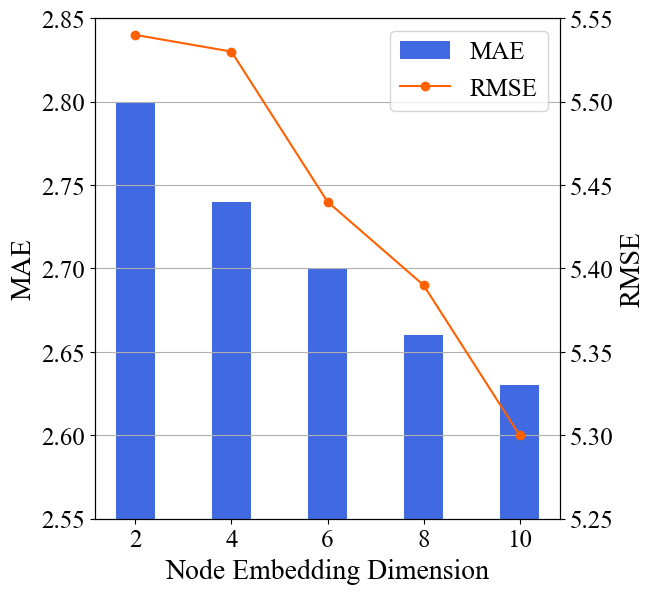}
		\caption{Effects of node embedding dimension ($D_{e}$) on \emph{PEMSD7(M)}}
	\end{subfigure}
	\caption{Hyperparameter experiments of \emph{I-ASTGCRN}}
	\label{fig6}
\end{figure*}

\begin{figure*}[t]
	\centering
	\begin{subfigure}{0.24\linewidth}
		\centering
		\includegraphics[width=1\linewidth]{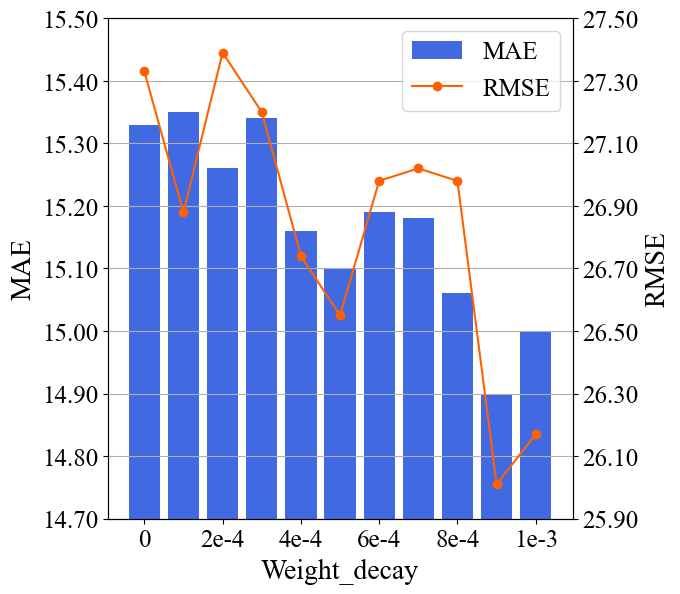}
		\caption{Effects of weight\_decay on \emph{PEMSD3}}
	\end{subfigure}
	\begin{subfigure}{0.24\linewidth}
		\centering
		\includegraphics[width=1\linewidth]{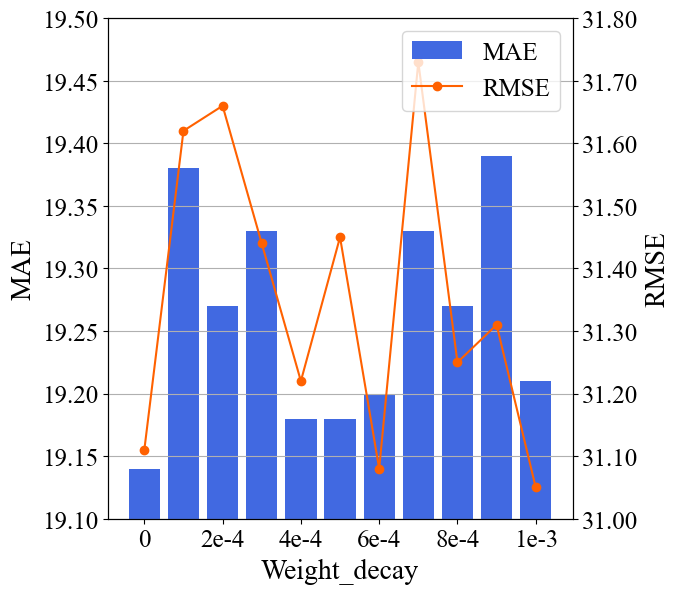}
		\caption{Effects of weight\_decay on \emph{PEMSD4}}
	\end{subfigure}
	\begin{subfigure}{0.24\linewidth}
		\centering
		\includegraphics[width=1\linewidth]{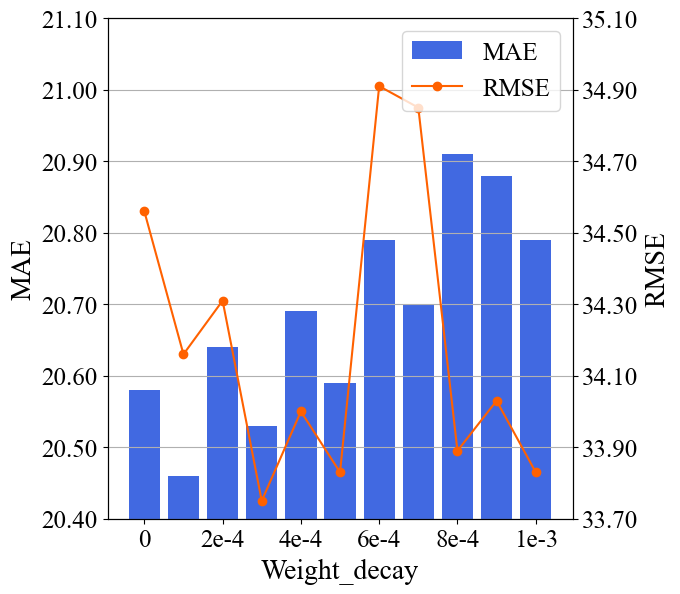}
		\caption{Effects of weight\_decay on \emph{PEMSD7}}
	\end{subfigure}
	\begin{subfigure}{0.24\linewidth}
		\centering
		\includegraphics[width=1\linewidth]{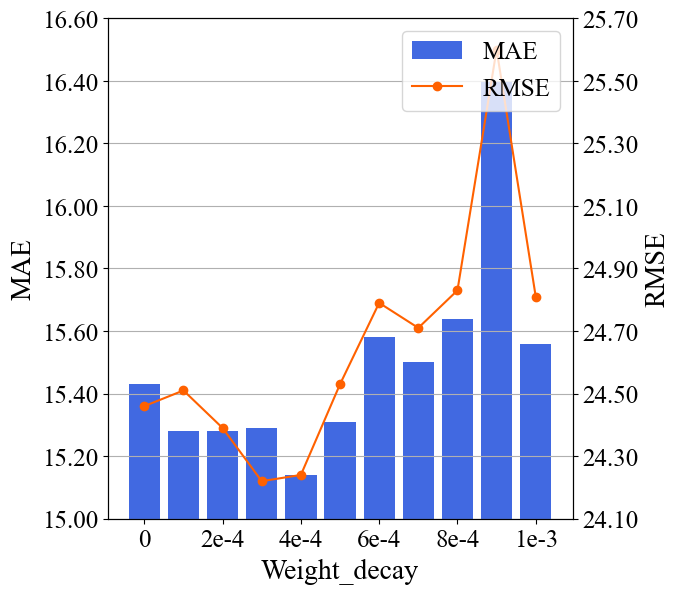}
		\caption{Effects of weight\_decay on \emph{PEMSD8}}
	\end{subfigure}

	\begin{subfigure}{0.24\linewidth}
		\centering
		\includegraphics[width=1\linewidth]{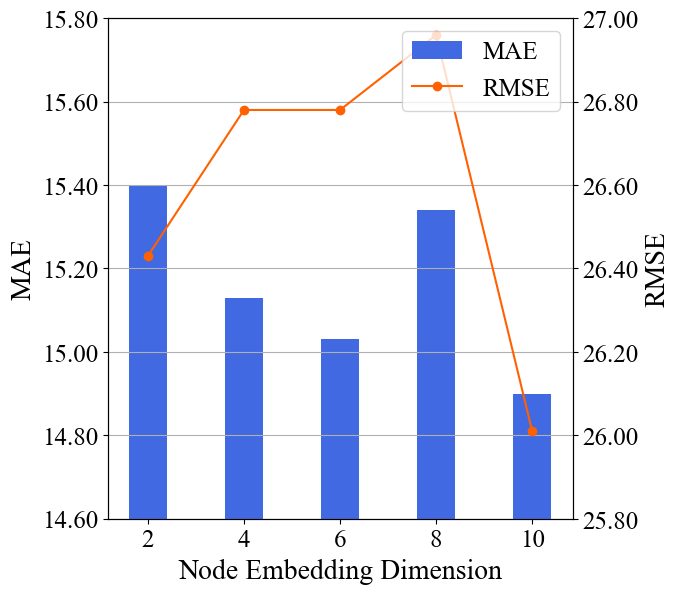}
		\caption{Effects of node embedding dimension ($D_{e}$) on \emph{PEMSD3}}
	\end{subfigure}
	\begin{subfigure}{0.24\linewidth}
		\centering
		\includegraphics[width=1\linewidth]{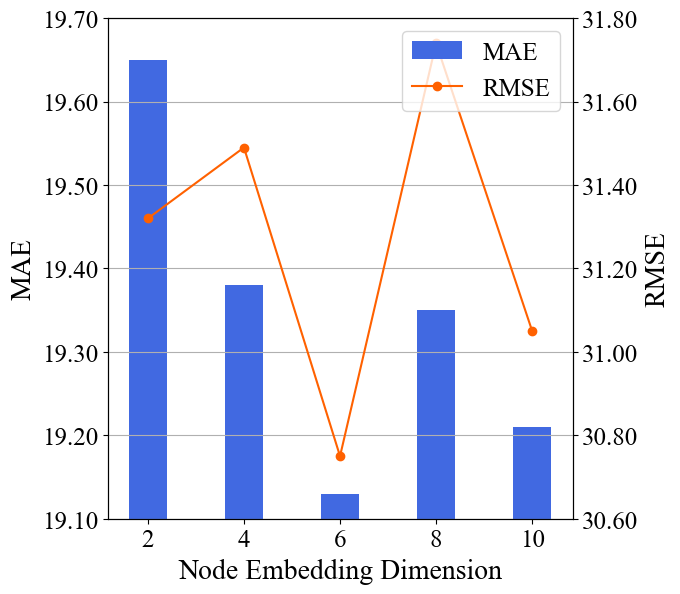}
		\caption{Effects of node embedding dimension ($D_{e}$) on \emph{PEMSD4}}
	\end{subfigure}
	\begin{subfigure}{0.24\linewidth}
		\centering
		\includegraphics[width=1\linewidth]{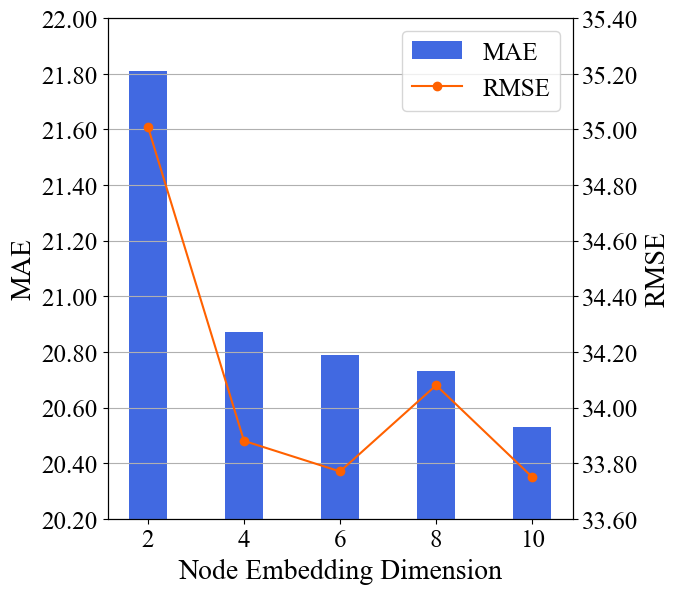}
		\caption{Effects of node embedding dimension ($D_{e}$) on \emph{PEMSD7}}
	\end{subfigure}
	\begin{subfigure}{0.24\linewidth}
		\centering
		\includegraphics[width=1\linewidth]{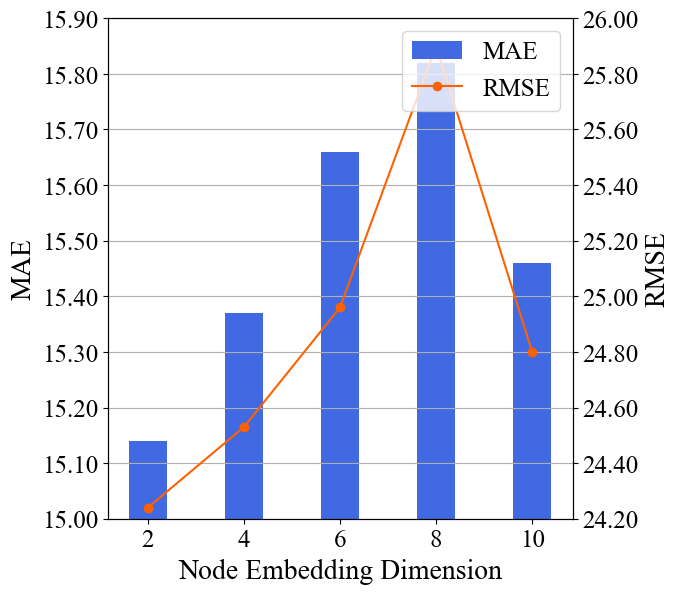}
		\caption{Effects of node embedding dimension ($D_{e}$) on \emph{PEMSD8}}
	\end{subfigure}
	\caption{Hyperparameter experiments of \emph{T-ASTGCRN}}
	\label{fig7}
\end{figure*}

\noindent Figures \ref{fig5}, \ref{fig6} and \ref{fig7} show the visualization results of hyperparametric experiments for \emph{A-STGCRN}, \emph{I-ASTGCRN} and \emph{T-ASTGCRN} that we did not report in the main text, respectively.

\end{document}